\documentclass{article}

\usepackage{microtype}
\usepackage{graphicx}
\usepackage{subcaption}
\usepackage{booktabs} 

\usepackage{hyperref}

\usepackage[accepted]{icml2026}

\usepackage{amsmath}
\usepackage{amssymb}
\usepackage{mathtools}
\usepackage{amsthm}

\usepackage[capitalize,noabbrev]{cleveref}

\theoremstyle{plain}
\newtheorem{theorem}{Theorem}[section]
\newtheorem{proposition}[theorem]{Proposition}

\theoremstyle{definition}

\theoremstyle{remark}

\usepackage[textsize=tiny]{todonotes}

\usepackage{hyperref}
\usepackage{url}
\usepackage[para]{footmisc}
\usepackage{comment}  

\usepackage{booktabs}     
\usepackage{multirow}     
\usepackage{multicol}     
\usepackage{wrapfig}       
\usepackage{subfiles}
\usepackage{bbm}

\usepackage{array}        

\usepackage{csquotes}
\usepackage[most]{tcolorbox}
\usepackage{caption}  

\usepackage{placeins}  

\usepackage{amsmath,amsfonts,bm}

\def\eqref#1{equation~\ref{#1}}

\def\1{\bm{1}}

\DeclareMathAlphabet{\mathsfit}{\encodingdefault}{\sfdefault}{m}{sl}
\SetMathAlphabet{\mathsfit}{bold}{\encodingdefault}{\sfdefault}{bx}{n}

\icmltitlerunning{Escaping the Verifier: Learning to Reason via Demonstrations}

\begin{document}

\twocolumn[
  \icmltitle{Escaping the Verifier: Learning to Reason via Demonstrations}



  \icmlsetsymbol{equal}{*}

  \begin{icmlauthorlist}
    \icmlauthor{Locke Cai}{together,mit}
    \icmlauthor{Max Ryabinin}{together}
    \icmlauthor{Ivan Provilkov}{together}
  \end{icmlauthorlist}

  \icmlaffiliation{mit}{Massachusetts Institute of Technology}
  \icmlaffiliation{together}{Together AI}

  \icmlcorrespondingauthor{Locke Cai}{lcai12@mit.edu}
  \icmlcorrespondingauthor{Ivan Provilkov}{ivan@together.ai}

  \icmlkeywords{Inverse Reinforcement Learning, Large Language Models}

  \vskip 0.3in
]



\printAffiliationsAndNotice{}  

\begin{abstract}

Training Large Language Models (LLMs) to reason often relies on Reinforcement Learning (RL) with task-specific verifiers. However, many real-world reasoning-intensive tasks lack verifiers, despite offering abundant expert demonstrations that remain under-utilized for reasoning-focused training. We introduce \textbf{RARO} (Relativistic Adversarial Reasoning Optimization), which learns strong reasoning capabilities from expert demonstrations alone via \textbf{Inverse Reinforcement Learning}. Our method sets up an adversarial game between a \textbf{policy} and a \textbf{relativistic critic}: the policy learns to mimic expert answers, while the critic aims to identify the experts among (expert, policy) answer pairs. Both the policy and the critic are trained jointly and continuously via RL, and we identify the key stabilization techniques required for robust learning. Empirically, RARO significantly outperforms strong verifier-free baselines across all evaluation tasks: $+13.7\%$ accuracy on Countdown ($1.5$B), +8.2\% on DeepMath ($7$B), and $+19.1\%$ win-rate on Poetry Writing ($7$B) against expert poems. RARO also exhibits similar robust scaling trends as RL with verifiers. These results demonstrate that RARO effectively elicits strong reasoning performance from expert demonstrations alone, enabling robust reasoning learning even when task-specific verifiers are unavailable.
\end{abstract}

\section{Introduction}

Recent advances in Large Language Models (LLMs) have been driven substantially by improvements in their \emph{reasoning} abilities. 
Reasoning enables LLMs to perform intermediate computations before answering user queries, proposing candidate solutions and self-corrections. 
Much of this progress has been enabled via Reinforcement Learning (RL) on \emph{verifiable} tasks such as mathematics and competitive programming~\citep{deepseek-r1,qwen3,grpo-deepseekmath}. 
Notably, recent work has demonstrated that RL with Verifiable Rewards (RLVR) can enable LLMs to develop robust reasoning capabilities without any additional supervision~\citep{deepseek-r1}. 
A growing body of work further improves the efficiency and stability of RL algorithms on verifiable tasks~\citep{yu2025dapoopensourcellmreinforcement,gspo}. 
Compared to that, little attention has been paid to the development of reasoning abilities on \emph{non-verifiable} tasks, i.e., tasks where verifiers do not exist. 

\begin{figure}[t]
    \begin{center}
    \includegraphics[width=\linewidth]{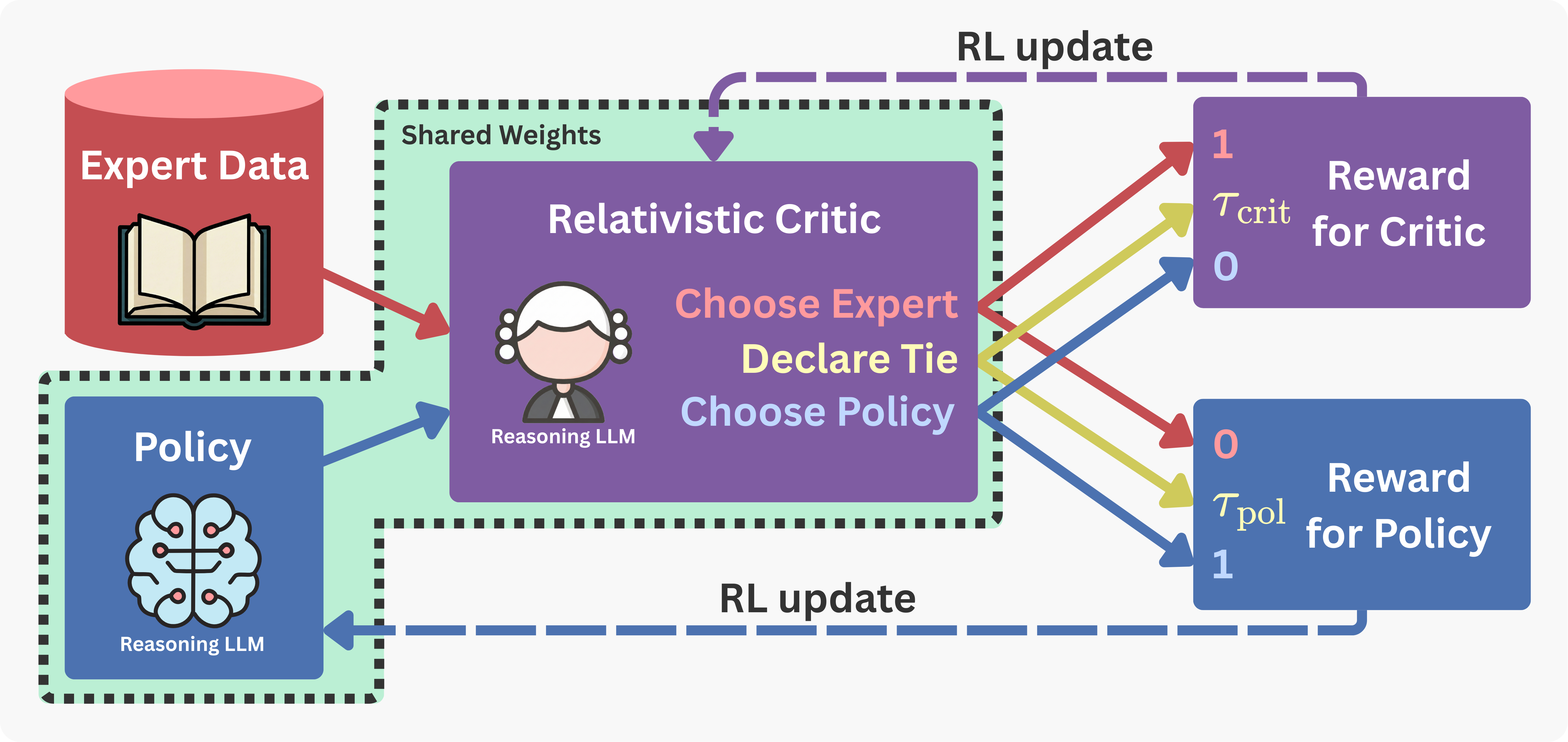}
    \end{center}
    \caption{\textbf{An overview of RARO.} The method creates an adversarial game between a \emph{policy} and a \emph{relativistic critic} that share the same weights. The critic is rewarded for identifying the expert responses among (expert, policy) answer pairs, while the policy is rewarded for deceiving the critic. Additionally, the critic can declare a \emph{tie}, yielding stable rewards when it is unsure. Both the policy and the critic are trained jointly via RL.}
    \label{fig:main_method}
\end{figure}

Yet, in many impactful and challenging tasks, such as analytical writing, open-ended research, or financial analysis, LLM outputs are not directly verifiable due to hard-to-specify criteria, wide variation among acceptable answers, and other constraints. 
A popular approach in these settings is Reinforcement Learning from Human Feedback (RLHF,~\citealp{instructgpt_NEURIPS2022_b1efde53, dpo}), but it requires collecting human preferences, which is often a time-consuming and expensive process.

Without preference data, the typical approach to improving LLM performance in these domains is to perform Supervised Fine-Tuning (SFT) on expert demonstrations. 
However, this method, even if the data are further annotated with reasoning traces, does not encourage the same reasoning behaviors as those arising from large-scale RL on verifiable tasks~\citep{chu2025sftmemorizesrlgeneralizes}. 
Additionally, naive use of next-token prediction induces a training-inference distribution mismatch: during training, the model conditions only on the dataset contexts, whereas at inference, it observes self-generated contexts. 
On-policy training, as occurs during RL, yields a lower training-inference mismatch, leading to better performance at test time~\citep{ross-learning-upper-bound}. 
Thus, we hypothesize that using expert demonstrations along with RL could offer a new pathway for developing robust reasoning capabilities in non-verifiable domains.

To this end, we introduce \textbf{RARO} (Relativistic Adversarial Reasoning Optimization), a robust RL algorithm that trains LLMs to reason using only expert demonstrations \emph{without} task-specific verifiers or human preferences.
RARO trains a single LLM to play two roles: a reasoning \emph{policy} that generates answers and a \emph{relativistic critic} that compares (expert, policy) answer pairs and predicts which is expert (or declares a \emph{tie}). It then alternates critic updates with policy updates using the critic as a learned reward signal.


The key contributions of our work are as follows:
\vspace{-1ex}

\begin{itemize}
    \item We propose a novel perspective on training reasoning models via \textbf{Inverse Reinforcement Learning}~\citep{ng2000algorithms}. With this perspective, we develop a principled method, RARO, that enables training reasoning models using demonstration data only. 
    \item We evaluate RARO on a controlled toy reasoning task, \textbf{Countdown}, where it not only significantly outperforms strong baselines without verification ($+13.7\%$ accuracy, Qwen2.5 $1.5$B), but it nearly matches the performance of RLVR, demonstrating the effectiveness of RARO on inducing reasoning behaviors.
    \item Next, we further stress test RARO's reasoning elicitation capability by scaling it on the general domain of math problems via the \textbf{DeepMath} dataset \citep{he2025deepmath103klargescalechallengingdecontaminated}, where RARO again outperforms baselines without verification ($+8.2\%$ accuracy, Qwen2.5 $7$B) and exhibits scaling trends similar to RLVR, demonstrating the \emph{scalability} of RARO.
    \item Finally, we show that RARO's superior performance generalizes to non-verifiable domains by evaluating it on \textbf{Poetry Writing}, where it considerably outperforms all baselines ($+19.1\%$ win rate, Qwen2.5 $7$B).
\end{itemize}

\section{Related Work}

\subsection{Reinforcement Learning with Verifiable Rewards}

Chain-of-Thought (CoT) prompting~\citep{wei2022chain} is a simple yet effective technique that enables LLMs to generate intermediate reasoning tokens, steering them toward correct answers. This approach has become a central focus for enhancing reasoning, pairing naturally with test-time scaling for further performance gains~\citep{snell2024scalingllmtesttimecompute}.

Currently, the standard method to elicit such strong CoT capabilities is Reinforcement Learning with Verifiable Rewards (RLVR). RLVR methods train LLMs to produce long reasoning traces on tasks with ground-truth verifiers, enabling recent open-source models to achieve expert-level performance~\citep{deepseek-r1, qwen3}. 

The dominant algorithm in this space is Group Relative Policy Optimization (GRPO,~\citealp{grpo-deepseekmath}), which builds upon Proximal Policy Optimization (PPO,~\citealp{ppo}) by estimating advantages via group-wise sample averages. 
Subsequent works like DAPO~\citep{yu2025dapoopensourcellmreinforcement} and GSPO~\citep{gspo} have further improved its efficiency and stability. 
However, while RLVR is highly effective, it is fundamentally constrained to verifiable tasks such as mathematics and competitive programming.

\subsection{General Reasoning Learning}

While RLVR is effective for training LLMs to reason on verifiable tasks, it does not directly extend to the broader setting of learning reasoning on real-world domains with no verifiers, yet many of these tasks could still benefit from explicit reasoning~\citep{zhou2025verifree}.

Although no consensus method exists for general reasoning learning, several recent efforts make early progress. \citet{zhou2025verifree} and \citet{gurung2025learningreasonlongformstory} propose to train LLMs to reason with reward derived from the model's own logits on expert answers. \citet{jia2025writingzerobridgegapnonverifiable} propose a pairwise generative reward model with a PPO-style objective for non-verifiable writing tasks. \citet{gunjal2025rubricsrewardsreinforcementlearning} propose using an LLM-as-judge~\citep{gu2025surveyllmasajudge} together with pre-generated rubrics from a strong LLM to provide rewards for non-verifiable tasks. \citet{ma2025generalreasoneradvancingllmreasoning} distill a model-based verifier from a strong teacher. \citet{tang2026beyond} extend RL
beyond verifiable rewards by assuming access to a known reference answer and rewarding the policy for matching it. \citet{li2025internbootcamptechnicalreportboosting} investigate large-scale multi-task RLVR, hypothesizing that breadth across many tasks induces stronger general reasoning. In contrast to these approaches, we adopt a demonstration-only setting and develop an inverse-RL perspective that trains reasoning models \emph{without} task verifiers and \emph{without} depending on a verifier-trained teacher.

\subsection{Inverse Reinforcement Learning}

Inverse Reinforcement Learning (IRL,~\citealp{ng2000algorithms}) studies the task of recovering a reward function for which an observed expert policy is near-optimal. A seminal application is robust imitation learning, most notably Generative Adversarial Imitation Learning (GAIL,~\citealp{ho2016generativeadversarialimitationlearning}), casting imitation as an adversarial game between a policy and a discriminator.

\citet{sun2025inverserlignmentlargelanguagemodel} recently investigated the application of IRL for aligning LLMs with expert demonstrations. 
They show that a classifier trained in the IRL paradigm can serve as an effective reward model for Best-of-N sampling. 
However, their work stops short of exploring stable, joint adversarial training, or reasoning-intensive tasks, where the model must learn to navigate complex solution spaces rather than aligning with surface-level preferences.
\section{Method} \label{sec:method}

We study the general setting where we are given an expert Question–Answer (QA) dataset, and we aim to train a reasoning LLM policy to produce expert-level answers via explicit CoT reasoning. We adopt this setting because verifiable tasks are relatively scarce, whereas expert demonstration data are abundant for many non-verifiable domains (e.g., highly upvoted Stack Exchange answers).

To approach this task, we propose a novel inverse reinforcement learning framework that sets up an adversarial interaction between a \emph{reasoning policy} and a \emph{relativistic critic}: the policy learns to output expert-like answers, while the critic learns to discriminate between policy and expert answers. By jointly training both the policy and the critic to reason via RL, we enable the emergence of strong reasoning capabilities from demonstrations alone, without requiring task-specific verifiers.

\subsection{Reasoning Optimization via Inverse RL}

Let \(D=\{(q_i,a_i)\}_{i=1}^n\) denote the expert QA dataset. 
We parameterize our reasoning policy as a \emph{conditional latent-variable model}~\citep{trice} $\pi_\theta(a,z \mid q)$,
a distribution over answers \(a\) and CoT reasoning traces \(z\) conditioned on a question \(q\). 
We let \(\hat{p}_q\) denote the empirical distribution of questions in \(D\), \(\hat{p}_{a\mid q}\) denote the empirical distribution of expert answers conditioned on a question, and \(\hat{p}_D = \hat{p}_{a\mid q} \hat{p}_q\) denote the empirical distribution of dataset pairs \((q,a)\).

A natural baseline for producing expert-quality answers is the \textit{maximum likelihood (ML)} objective on expert demonstrations: \(\arg\max_{\theta}\; \mathbb{E}_{(q,a)\sim \hat{p}_D}\!\left[\log \pi_{\theta}(a \mid q)\right]\).

However, for models that perform CoT reasoning before producing an answer, each \((q,a)\) is associated with many possible CoT traces. Thus, the marginal likelihood required by the ML objective, \(\pi_{\theta}(a \mid q) = \sum_{z} \pi_{\theta}(a,z \mid q)\), involves summing over a combinatorially large (often effectively unbounded) set of traces, rendering exact computation and its gradients computationally impractical.

To address this intractability, we adopt the perspective of Inverse Reinforcement Learning (IRL). 
Rather than maximizing the marginal likelihood directly, we learn a \emph{parameterized reward} \(r_\phi(a,q)\) over QA pairs such that optimizing \(\pi_\theta\) with respect to \(r_\phi\) yields a \enquote{near-optimal} policy that approximately maximizes the ML objective.

We formalize \enquote{near-optimality} via the KL-regularized reward-maximization objective. 
Under this objective, it can be shown \citep{awr} that the optimal policy has the following marginal distribution:
\begin{equation}
\!\!\pi_{\theta^\star(\phi)}(a\! \mid\! q) 
\!=\! \frac{1}{Z_{\theta^\star(\phi)}(q)} \, 
\pi_{\mathrm{ref}}(a\! \mid\! q) \,
\!\exp\!\left\{\frac{1}{\beta} \, r_\phi(a,q)\right\}\!,
\end{equation}
where \(Z_{\theta^\star(\phi)}(q)\) is the partition function, \(\pi_{\mathrm{ref}}\) is a fixed reference policy, and \(\beta>0\) controls the strength of the KL-regularization. See Appendix \ref{apn:derivations:closed-form-policy} for the full proof.

With a closed-form expression for the optimal marginal under reward model \(r_\phi\), we have a concrete pathway to optimize \(\phi\) such that \(\pi_{\theta^\star(\phi)}\) maximizes the ML objective. Specifically, as shown in Appendix \ref{apn:derivations:reward-grad}, we can directly optimize \(\phi\) via gradient ascent with the following gradient:
\begin{gather}
\begin{aligned}
\nabla_\phi \mathcal{L}(\phi)
&\!=\! \frac{1}{\beta}\Big(
\underbrace{\mathbb{E}_{(q,a)\sim \hat{p}_D}\!\big[\nabla_\phi r_\phi(a,q)\big]}_{\text{expert answers}} \\
&-\underbrace{\mathbb{E}_{q\sim \hat{p}_q}\,
\mathbb{E}_{a'\sim \pi_{\theta^\star(\phi)}(\cdot\mid q)}\!\left[\nabla_\phi r_\phi(a',q)\right]}_{\text{policy answers}}
\Big).
\end{aligned}
\end{gather}
Finally, approximating \(\pi_{\theta^\star(\phi)}\) by optimizing \(\pi_\theta\) with reward \(r_\phi\) via RL, we have an algorithm that optimizes the reasoning policy \(\pi_\theta\) with respect to the maximum likelihood objective, as shown in Algorithm \ref{alg:alt-policy-reward-opt}.

\begin{algorithm}[t]
  \caption{Alternating Policy-Reward Optimization}
  \label{alg:alt-policy-reward-opt}
  \textbf{Inputs:} Dataset $D=\{(q_i,a_i)\}$; Batch $B$; Learning rates $\eta_r,\eta_\pi$.\\
  \textbf{Models:} Reward $r_\phi(a,q)$; Reasoning policy $\pi_\theta(a,z\mid q)$.
  \begin{algorithmic}
  \STATE Initialize $\phi,\theta$
  \FOR{$t=1,\ldots,T$}
    \STATE Draw $\{(q_i,a_i^{\mathrm{E}})\}_{i=1}^{B}$ with $q_i\!\sim\!\hat p_q,\; a_i^{\mathrm{E}}\!\sim\!\hat p_{a\mid q}(\cdot\mid q_i)$
    \STATE For each $i\in[1..B]$, sample $(z_{i},a^{\mathrm{P}}_{i})\!\sim\!\pi_\theta(\cdot,\cdot\mid q_i)$
    \STATE \textbf{Reward update:}
    \[
    \phi \leftarrow \phi + \eta_r \cdot \frac{1}{\beta B}
    \sum_{i=1}^{B}
    \Big(
    \nabla_\phi r_\phi(a_i^{\mathrm{E}}, q_i)
    -
    \nabla_\phi r_\phi(a_i^{\mathrm{P}}, q_i)
    \Big)
    \]
    \STATE \textbf{Policy update:} KL-regularized GRPO with $r_\phi(a,q)$ as the reward.
  \ENDFOR
  \end{algorithmic}
\end{algorithm}

\subsection{Reasoning Critic as Reward Model}

To instantiate this framework, we need to decide on an appropriate architecture for the reward model \(r_\phi\). Our setting targets difficult QA tasks that benefit from reasoning. Thus, to separate expert from policy answers, we expect the reward model to be at least as capable as the policy. 

To this end, we represent \(r_\phi\) with a reasoning \emph{critic} \(c_{\phi}\) that takes an \((q, a)\) pair and outputs \(\ell \in \{\mathrm{expert}, \mathrm{policy}\}\), classifying whether an answer is from the expert or the policy. Specifically, we parametrize \(r_\phi\) as the probability of classifying as \emph{expert} for a \((q, a)\) pair.

Under this parameterization, as shown in Appendix \ref{apn:derivations:reasoning-reward}, the gradient \(\nabla_{\phi}L\) corresponds to the standard \emph{policy gradient}, resulting in simple reward functions for the critic and policy.

\paragraph{Reward for Critic:} 
\begin{equation}
R_{\text{critic}}(\ell, a, q) = \mathbbm{1}_{\ell \text{ is correct}}\ ,
\end{equation}

\paragraph{Reward for Policy:}
\begin{equation}
R_{\text{policy}}(a, q) = \mathbbm{1}_{\ell = \mathrm{expert}}, \quad \ell \sim c_{\phi}(\cdot \mid a, q).
\end{equation}

This allows us to optimize both the critic and the policy using the same GRPO algorithm. 

Such reward formulation creates an \emph{adversarial game} between the critic and policy: the critic is rewarded when it correctly classifies the answer as coming from the expert or policy, while the policy is rewarded when the critic \emph{incorrectly} classifies its answer as an expert answer.


\subsection{Relativistic Critic: The Pairwise Advantage} \label{sec:method:relativistic-critic}

Despite the theoretical soundness, the previous setup poses challenges for critic learning. As policy approaches the expert the classification task becomes much more difficult due to a lack of reference answer for the critic to compare against. In addition, with an optimal policy, the critic effectively degenerates to random guessing, providing high-variance, uninformative gradients to the policy, leading to training instability as we observed (see Appendix \ref{sec:ablation_studies}).

To address these limitations, we adopt a \emph{relativistic} formulation in which the critic compares a policy answer against an expert answer for the same question and outputs which is better, or a \texttt{tie} if they are of equal quality. This avoids degeneracy when the policy is optimal. We empirically show that allowing \texttt{tie} is crucial for performance, reporting the details of this experiment in Appendix~\ref{sec:ablation_studies}.

Formally, the \emph{relativistic critic} \(c_{\phi}\) takes a question \(q\) and two candidate answers \((a^{(1)}, a^{(2)})\) and returns a label \(\ell \in \{\text{1},\text{2},\text{tie}\}\). Assuming one expert and one policy answer, we can define: 

\paragraph{Reward for Critic:} 
\begin{equation}
\begin{aligned}
R_{\text{critic}}\bigl(q, a^{(1)}, a^{(2)}\bigr) &= \mathbbm{1}_{\ell \text{ is expert}} + \tau_{\text{crit}} \cdot \mathbbm{1}_{\ell = \text{tie}},\\
&\tau_{\text{crit}} \in [0,1],
\end{aligned}
\end{equation}

\paragraph{Reward for Policy:} 
\begin{equation}
\begin{aligned}
R_{\text{policy}}\bigl(q, a^{(1)}, a^{(2)}\bigr) &= \mathbbm{1}_{\ell \text{ is policy}} + \tau_{\text{pol}} \cdot \mathbbm{1}_{\ell = \text{tie}},\\
&\tau_{\text{pol}} \in [0,1],
\end{aligned}
\end{equation}

where \(\tau_{\text{crit}}\) and \(\tau_{\text{pol}}\) are \emph{tie rewards} --- hyperparameters introduced to handle the \texttt{tie} label.

Unlike the binary classification setup, the relativistic critic is now given a \emph{pairwise comparison} task: the critic is rewarded when it correctly identifies the expert answer, and the policy is rewarded when the critic mistakenly identifies its answer as the expert answer, with additional \texttt{tie} rewards to ensure non-degeneracy and stable learning. 

\subsection{RARO: Relativistic Adversarial Reasoning Optimization}

Finally, we identify the key techniques for stable and efficient training. First, we use a \emph{shared LLM} for both the critic and the policy, which reduces memory usage and promotes generalization. This allows us to employ \emph{data mixing}, where policy and critic rollouts are combined in a single batch, simplifying the training loop. To prevent the critic from catastrophic forgetting, we sample critic prompts from a \emph{replay buffer} that stores all past expert and policy answers. Finally, we incorporate several practical improvements to the GRPO algorithm, such as over-length filtering and removing advantage/length normalization. For full implementation details, please refer to Appendix~\ref{sec:implementation_details_stable_learning}.

Incorporating all of these optimizations, we arrive at our final algorithm, \emph{RARO (Relativistic Adversarial Reasoning Optimization)}, shown in Algorithm~\ref{alg:unified}.

\begin{algorithm}[t]
  \caption{RARO (Relativistic Adversarial Reasoning Optimization)}
  \label{alg:unified}

  \textbf{Inputs:} Dataset $D=\{(q_i,a_i)\}$; tie rewards $\tau_{\mathrm{pol}},\tau_{\mathrm{crit}}$;
  loss weights $\lambda_{\mathrm{pol}},\lambda_{\mathrm{crit}}$; batch $B$; rollout $K$.\\
  \textbf{Model:} Shared $\theta \to \pi_\theta, c_\theta$; replay buffer $\mathcal{R}$.

  \begin{algorithmic}[1]
  \STATE Initialize $\theta,\ \mathcal{R}\gets\emptyset$
  \FOR{$t=1,\ldots,T$}
    \STATE $\mathcal{R}_{\text{new}} \gets \emptyset$
    \STATE Draw $\{(q_i, a^{\mathrm{E}}_i)\}_{i=1}^B \sim D$
    \STATE \textbf{Policy rollouts:}
    \FOR{$i=1,\ldots,B,\ k=1,\ldots,K$}
      \STATE $(z^{\mathrm{P}}_{i,k}, a^{\mathrm{P}}_{i,k}) \sim \pi_\theta(\cdot \mid q_i)$, build pair $(a^{\mathrm{E}}_i, a^{\mathrm{P}}_{i,k})$
      \STATE $\ell_{i,k} \sim c_\theta(\cdot \mid q_i, a^{\mathrm{E}}_i, a^{\mathrm{P}}_{i,k})$
      \STATE $R^{\mathrm{pol}}_{i,k} \gets
        \mathbbm{1}_{\ell_{i,k}\ \text{is policy}}
        + \tau_{\mathrm{pol}}\mathbbm{1}_{\ell_{i,k}=\text{tie}}$
      \STATE $\mathcal{R}_{\text{new}} \gets
        \mathcal{R}_{\text{new}} \cup \{(q_i, a^{\mathrm{E}}_i, a^{\mathrm{P}}_{i,k})\}$
    \ENDFOR
    \STATE $\mathcal{C} \gets \text{Mix}(\mathcal{R}_{\text{new}}, \mathcal{R})$
    \STATE $\mathcal{R} \gets \mathcal{R} \cup \mathcal{R}_{\text{new}}$
    \STATE \textbf{Critic rollouts:}
    \FOR{$(q_j, a^{\mathrm{E}}_j, a^{\mathrm{P}}_j)\in\mathcal{C}$}
      \STATE $\ell_j \sim c_\theta(\cdot \mid q_j, a^{\mathrm{E}}_j, a^{\mathrm{P}}_j)$
      \STATE $R^{\mathrm{crit}}_j \gets
        \mathbbm{1}_{\ell_j\ \text{is expert}}
        + \tau_{\mathrm{crit}}\mathbbm{1}_{\ell_j=\text{tie}}$
    \ENDFOR

    \STATE GRPO step on $\theta$ to maximize:
    $\lambda_{\mathrm{pol}} J_{\mathrm{pol}}(\theta)
    + \lambda_{\mathrm{crit}} J_{\mathrm{crit}}(\theta)
    - \beta D_{\mathrm{KL}}(\pi_\theta \| \pi_{\mathrm{ref}})$
  \ENDFOR
  \STATE \textbf{return} $\theta$
  \end{algorithmic}
\end{algorithm}
  \vspace{-4pt}

\section{Experimental Setup} \label{sec:experiments}

\subsection{Tasks \& Datasets}

We evaluate RARO on three diverse reasoning tasks that cover complementary aspects of reasoning.
See Appendix~\ref{sec:dataset_details} for more details on the datasets.

\paragraph{Countdown.}
First, we evaluate our method on the Countdown task, a controlled toy reasoning task where answer verification is much simpler than answer generation. We use a 24-style variant where the goal is to combine four integers to obtain 24 (see Appendix~\ref{sec:dataset_details} for details). Through this task, we aim to study the effectiveness of our method on reasoning capabilities in a controlled environment where answer checking is much easier than solution search.

\paragraph{DeepMath.}
Next, we evaluate our method on general mathematical reasoning using the DeepMath dataset~\citep{he2025deepmath103klargescalechallengingdecontaminated}. Compared to Countdown, answer verification in this domain is significantly harder, often requiring solving the problem from scratch. Through this task, we aim to stress-test our method in complex reasoning environments, where verification is as difficult as generation.

\vspace{-6pt}

\paragraph{Poetry Writing.}
Finally, we extend our method to its intended setting of non-verifiable, open-ended reasoning tasks. While there exist benchmarks for non-verifiable domains~\citep{arora2025healthbenchevaluatinglargelanguage, paech2024eqbenchemotionalintelligencebenchmark}, there is no official training data associated with them. Thus, we choose to evaluate our method and baselines on a custom Poetry Writing dataset. 
Unlike math tasks, poetry writing does not admit an objective verifier. 
Thus, for evaluation, we use GPT-5 ~\citep{openai2025gpt5} as a judge to evaluate poems both in isolation and in comparison to the expert poem (see Appendix~\ref{sec:dataset_details} for details). 
This task represents the non-verifiable regime that our method aims to handle, where explicit reasoning could significantly improve quality.

\vspace{-0.5ex}
\subsection{Baselines}
\vspace{-0.5ex}

We compare RARO against strong post-training baselines under the same data, training, and evaluation setup.

\vspace{-0.5ex}
\paragraph{Supervised Fine-Tuning (SFT).}
The SFT baseline trains the base models to directly maximize the conditional log-likelihood of the expert answer given the question, representing the standard use of demonstration data.

\paragraph{Rationalization.}
Following prior work on self-rationalizing techniques~\citep{zelikman2022starbootstrappingreasoningreasoning}, we construct a rationalization baseline that augments each expert answer with an explicit CoT. Concretely, we prompt the base model to annotate the expert demonstrations with free-form \emph{rationale}, then perform SFT on the concatenated (question, rationale, answer) sequences, attempting to incentivize the base model to learn to reason. 

\paragraph{Iterative DPO.}
Direct Preference Optimization (DPO, \citealp{dpo}) is a natural way to match the policy's output distribution to the expert. 
Inspired by Iterative Reasoning Preference Optimization~\citep{pang2024iterativereasoningpreferenceoptimization}, we perform 3 rounds of DPO: in each round, we sample one response per question to form preference pairs favoring the expert. 
We initialize from the SFT checkpoint to prevent distribution mismatch and report the best result across rounds. 

\paragraph{RL from logit-based reward (RL-Logit).}
Recent work has proposed training reasoning LLMs via RL where the reward is derived from the model's own logits on expert answers rather than from an external verifier~\citep{zhou2025verifree,gurung2025learningreasonlongformstory}. We implement two variants of such \emph{logit-based} rewards, and report the metrics from the best performing variant (see Appendix~\ref{sec:implementation_details_stack} for details): 
\begin{itemize}
    \item The \emph{log-probability reward}, which uses the log-probability of the expert answer $a^\star$ given the question $q$ and generated reasoning tokens $z$ as the scalar reward $\log \pi_\theta(a^\star \mid q, z)$; 
    \item The \emph{perplexity reward}, which instead maximizes the negative perplexity of the expert answer under the same conditional distribution.
\end{itemize}

\paragraph{RL with Verifiable Reward (RLVR).}
For Countdown and DeepMath, where ground-truth verifiers are available, we additionally include a RLVR baseline trained with GRPO on binary rewards given by the verifier. This corresponds to the standard RLVR setting, and serves as an upper-bound for our method on tasks where verification is accessible.

\subsection{Training \& Evaluation Setup}

We evaluate our method and baselines on the Qwen2.5 \citep{qwen2025qwen25technicalreport} family of models. To focus on improving reasoning performance rather than language understanding, we initialize from instruction-tuned checkpoints instead of base checkpoints. 
We select the Qwen2.5 family because they are \emph{non-reasoning} LLMs, allowing us to study the effectiveness of our method on eliciting reasoning behaviors in a controlled manner.

We evaluate on Countdown and DeepMath with a ground-truth verifier, while Poetry writing is scored with GPT-5 as a judge in two ways: the \emph{scalar score} normalized from 1--7 to 0--100 and the \emph{win rate} against the expert poem. 
In both cases, we prompt GPT-5 to focus on prompt adherence and literary qualities. 
See Appendix~\ref{sec:dataset_details} for details.

Each dataset is split into training, validation, and test sets. 
For each method, we select the checkpoint with the highest validation performance and report the standard deviations computed via $100$ bootstrap repeats on the test set.
To ensure a fair comparison, we match dataset splits, rollout budgets, hyperparameters, and sampling configurations for all methods when possible. 
Unless otherwise specified, all methods are trained and evaluated with a reasoning budget of $2048$ tokens.
Appendix~\ref{sec:implementation_details} provides full implementation details and hyperparameters.

\vspace{-1ex}
\section{Main Results} \label{sec:results}

We present our experimental results organized by task: Countdown, DeepMath, and Poetry Writing. Across these domains, we observe that our method significantly and consistently outperforms all baselines, scaling effectively with both the reasoning budget and the model size.

\subsection{Countdown}

\begin{table}[t]
\centering
\caption{\textbf{Main Countdown Results.} RARO against baselines at a fixed reasoning budget of 2048 tokens.}
\begin{tabular}{lc}
\toprule
Method & Countdown accuracy (\%) $\uparrow$ \\
\midrule
RLVR (\emph{with verifier}) & $57.7 \pm 1.6$ \\
\midrule
Base & $2.0 \pm 0.4$ \\
SFT & $40.7 \pm 1.6$ \\
Rationalization & $12.5 \pm 1.0$ \\
Iterative DPO & $40.4 \pm 1.5$ \\
RL-Logit & $2.2 \pm 0.4$ \\
\textbf{RARO} & $\mathbf{54.4 \pm 1.5}$ \\
\bottomrule
\end{tabular}
\label{tab:countdown_main}
\vspace{1ex}
\end{table}
    
\begin{figure}[t]
\centering
\includegraphics[width=\columnwidth]{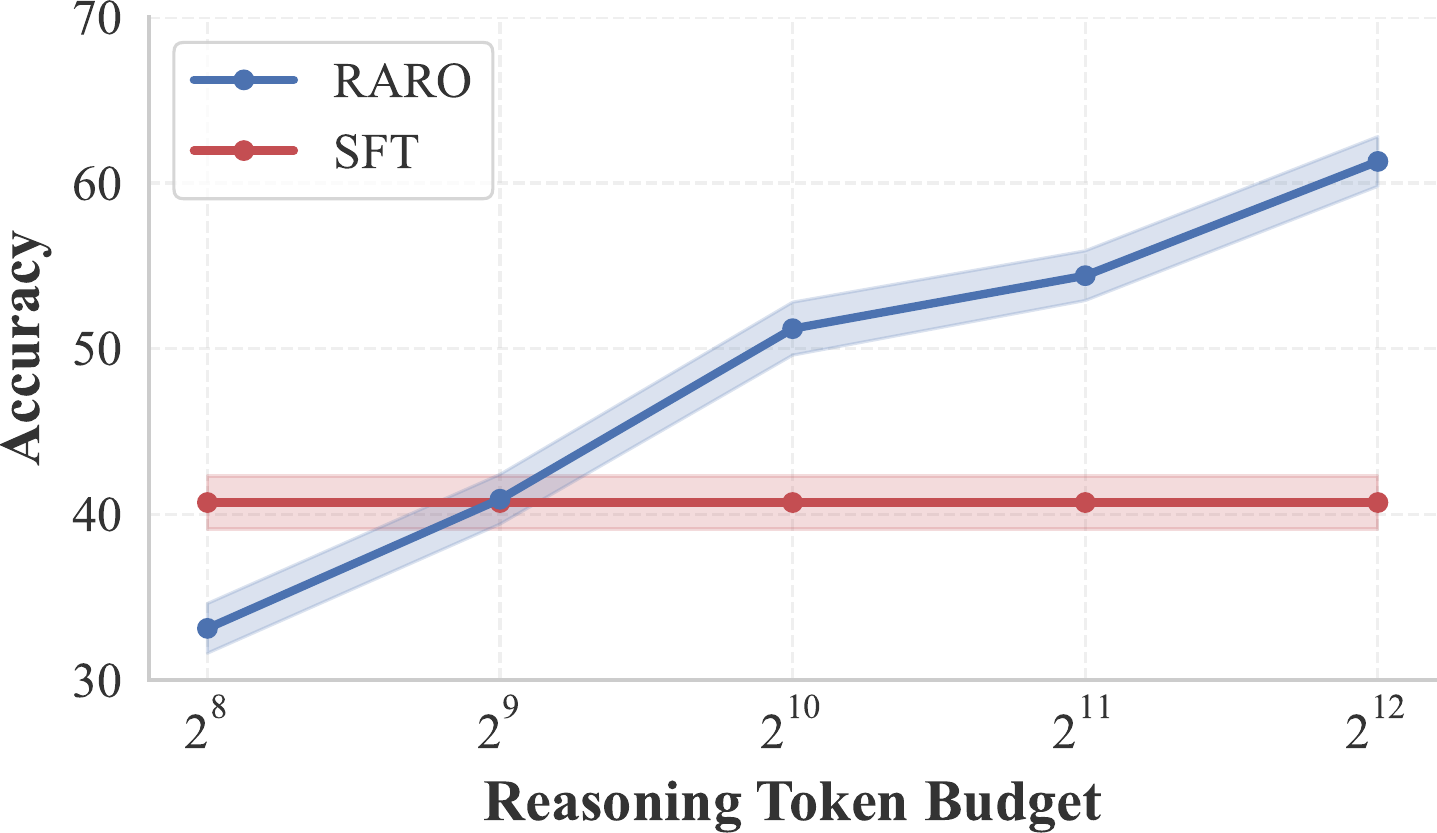}
\caption{\textbf{Reasoning Budget Scaling on Countdown.}
RARO scales effectively with both training and test-time token budgets.
See Appendix~\ref{sec:additional_tables_figures} for details.}
\label{fig:budget_scale}
\end{figure}

We first evaluate RARO on the Countdown task, a controlled toy reasoning task where answer verification is much simpler than answer generation. 
For this task, we focus our investigation on the $1.5$B model size and further ablate our method and baselines with respect to both training and test-time reasoning token budgets. 
We do not study the effect of model size, as Countdown is a task where the reasoning budget is the primary bottleneck rather than model capacity (see Appendix~\ref{sec:countdown_scaling} for details).

\vspace{-0.5ex}
\paragraph{Superior Performance at Fixed Budget.}
As shown in Table~\ref{tab:countdown_main}, at a fixed reasoning budget of 2048 tokens, RARO achieves the accuracy of $54.4\%$, significantly outperforming the best verifier-free baseline (SFT, $40.7\%$) by $13.7\%$ and nearly matching the oracle RLVR baseline ($57.7\%$). We also notice that RL-Logit ($2.2\%$) and Rationalization ($12.5\%$) perform rather poorly, and we hypothesize that it is likely due to the base model's inability to produce high-quality rationalizations or informative logits. The strong performance of RARO demonstrates that our learned critic provides signal that is comparable to verification rewards.

\vspace{-0.5ex}
\paragraph{Emergence of Self-Correcting Search.}
Our key qualitative finding is the emergence of explicit search behavior. 
As shown in Appendix~\ref{sec:additional_tables_figures}, our model learns to explore the solution space, dynamically proposing combinations, verifying them, and backtracking when they are incorrect (e.g., \enquote{too high}), which is absent in the SFT baseline.
This self-correction mechanism acts as an internal verifier, allowing the model to recover from errors.

\vspace{-0.5ex}
\paragraph{Scaling with Reasoning Budget.}
Finally, we examine the scalability of RARO with respect to both training and test-time reasoning token budget. Figure~\ref{fig:budget_scale} illustrates a clear trend: while the SFT baseline's performance plateaus at $40.7\%$ regardless of the token budget, our method exhibits continuous improvement as the budget increases, rising from $33.1\%$ at 256 tokens to $61.3\%$ at 4096 tokens. Notably, the result at 4096 tokens is achieved by a model trained with a 2048-token budget, demonstrating that our method can extrapolate to longer reasoning chains at test time without additional training. This scaling behavior confirms that RARO successfully transforms reasoning budget into better performance, a hallmark of effective reasoning.

\subsection{DeepMath}
Next, we evaluate RARO on the DeepMath dataset, a collection of general mathematical problems. 
For this task, we focus on scaling our method and baselines with respect to model size instead of the reasoning budget, as it is a much more difficult setting where model capacity is the real bottleneck in performance.

\begin{table*}[t]
    \caption{\textbf{Main results for DeepMath and Poetry.} All methods have a reasoning budget of 2048 tokens. For Iterative DPO, we report the maximum over 3 rounds. For RL-Logit, we report the best over 2 variants. See Table~\ref{tab:main_results_all_appendix} in Appendix~\ref{sec:additional_tables_figures} for full data.}
    \label{tab:main_results}
    \setlength{\tabcolsep}{12pt}
    \centering
    \begin{tabular}{c c c c c}
    \toprule
    \multirow{2}{*}{Model size} & 
    \multirow{2}{*}{Method} &
    \multicolumn{1}{c}{DeepMath} &
    \multicolumn{1}{c}{Poetry} &
    \multicolumn{1}{c}{Poetry} \\
    & & accuracy ($\%$) $\uparrow$ & score (0--100) $\uparrow$ &
    \begin{tabular}[t]{@{}c@{}}win rate ($\%$) $\uparrow$
    \end{tabular}
    \\
    \midrule

    \multirow{7}{*}{\textbf{1.5B}} &
    \quad RLVR (\emph{with verifier}) & $50.9 \pm 1.9$ & \textsc{N/A} & \textsc{N/A} \\
    \cmidrule(l){2-5}
    & \quad Base & $29.6 \pm 1.9$ & $35.0 \pm 0.9$ & $0.0 \pm 0.0$ \\
    & \quad SFT & $35.7 \pm 1.8$ & $53.7 \pm 1.0$ & $2.3 \pm 1.0$ \\
    & \quad Rationalization & $34.5 \pm 2.0$ & $35.6 \pm 1.6$ & $0.8 \pm 0.5$ \\
    & \quad Iterative DPO & $33.0 \pm 1.9$ & $48.6 \pm 0.9$ & $0.0 \pm 0.0$ \\
    & \quad RL-Logit & $37.7 \pm 1.9$ & $36.4 \pm 0.7$ & $0.0 \pm 0.0$ \\
    & \quad \textbf{RARO} & $\mathbf{41.3 \pm 1.9}$ & $\mathbf{67.8 \pm 0.8}$ & $\mathbf{7.8 \pm 1.7}$ \\
    \midrule

    \multirow{7}{*}{\textbf{3B}} &
    \quad RLVR (\emph{with verifier}) & $55.8 \pm 2.0$ & \textsc{N/A} & \textsc{N/A} \\
    \cmidrule(l){2-5}
    & \quad Base & $39.4 \pm 1.9$ & $46.5 \pm 0.9$ & $0.0 \pm 0.0$ \\
    & \quad SFT & $39.0 \pm 1.9$ & $57.4 \pm 1.0$ & $2.3 \pm 1.0$ \\
    & \quad Rationalization & $32.3 \pm 1.9$ & $30.8 \pm 1.9$ & $0.4 \pm 0.4$ \\
    & \quad Iterative DPO & $34.2 \pm 1.9$ & $69.8 \pm 0.8$ & $6.6 \pm 1.5$ \\
    & \quad RL-Logit & $43.1 \pm 2.0$ & $46.9 \pm 0.8$ & $0.4 \pm 0.4$ \\
    & \quad \textbf{RARO} & $\mathbf{49.1 \pm 1.9}$ & $\mathbf{71.9 \pm 0.8}$ & $\mathbf{17.2 \pm 2.4}$ \\
    \midrule

    \multirow{7}{*}{\textbf{7B}} &
    \quad RLVR (\emph{with verifier}) & $66.2 \pm 1.9$ & \textsc{N/A} & \textsc{N/A} \\
    \cmidrule(l){2-5}
    & \quad Base & $44.2 \pm 2.1$ & $54.0 \pm 0.9$ & $1.2 \pm 0.7$ \\
    & \quad SFT & $42.3 \pm 1.9$ & $65.4 \pm 1.0$ & $5.9 \pm 1.4$ \\
    & \quad Rationalization & $48.6 \pm 1.9$ & $57.7 \pm 1.2$ & $5.1 \pm 1.3$ \\
    & \quad Iterative DPO & $36.9 \pm 2.0$ & $66.5 \pm 0.9$ & $5.1 \pm 1.4$ \\
    & \quad RL-Logit & $49.3 \pm 2.0$ & $55.4 \pm 0.8$ & $3.9 \pm 1.2$ \\
    & \quad \textbf{RARO} & $\mathbf{57.5 \pm 2.0}$ & $\mathbf{77.3 \pm 0.8}$ & $\mathbf{25.0 \pm 2.6}$ \\
    \bottomrule
    \end{tabular}
\end{table*}

\vspace{-0.5ex}
\paragraph{Significant Improvement over Baselines.}
As we show in Table~\ref{tab:main_results}, RARO consistently outperforms all verifier-free baselines across model scales. 
With the 1.5B model, we achieve $41.3\%$ accuracy compared to $37.7\%$ for the best baseline (RL-Logit), an improvement of $3.6\%$. 
This advantage grows with model size: at 3B, our method ($49.1\%$) surpasses the best baseline (RL-Logit, $43.1\%$) by $6.0\%$, and at 7B, it reaches $57.5\%$, beating the baseline by $8.2\%$. 
These results demonstrate that RARO provides a strong signal for reasoning that outperforms not only purely supervised approaches like SFT or Rationalization, but also RL-Logit.

\vspace{-0.5ex}
\paragraph{Stable Training Dynamics.}
We further analyze the training dynamics of RARO on DeepMath. As shown in Figure~\ref{fig:deepmath_length_reward} and further in Appendix~\ref{sec:future_work}, our coupled training objective maintains a robust equilibrium, allowing the policy to steadily improve its reasoning capabilities and response length without collapsing. 
This stability confirms the robustness of our optimization procedure.

\begin{figure}[t]
        \centering
        \includegraphics[width=\linewidth]{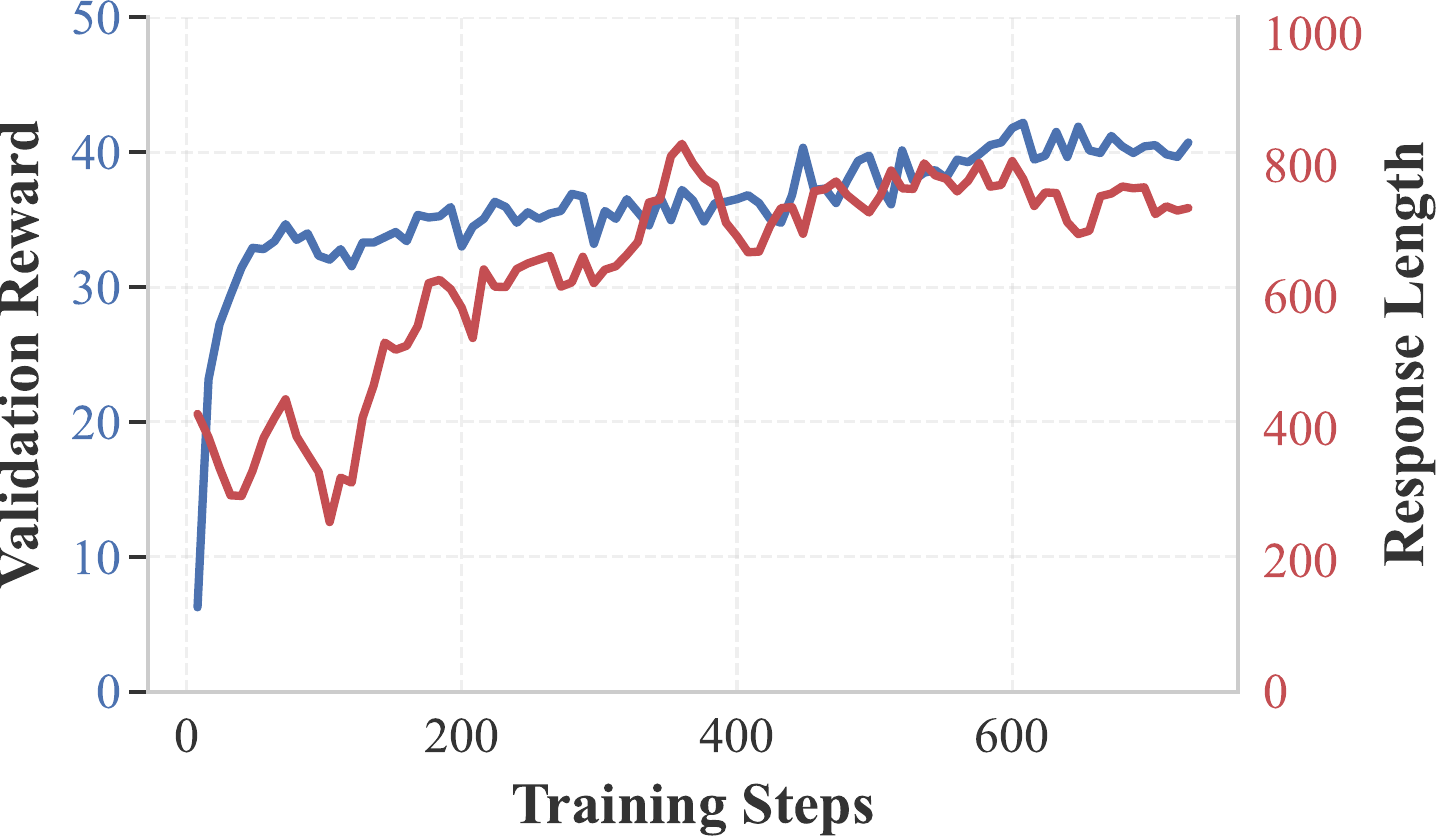}
        \caption{\textbf{Stable Reward and Length Growth.} The validation reward and response length of RARO on DeepMath (1.5B) continuously grows over time, indicating a stable dynamic.}
        \label{fig:deepmath_length_reward}
\end{figure}

\vspace{-0.5ex}
\paragraph{Effective Test-Time Scaling.}

\begin{figure}[t]
  \centering
  \includegraphics[width=\columnwidth]{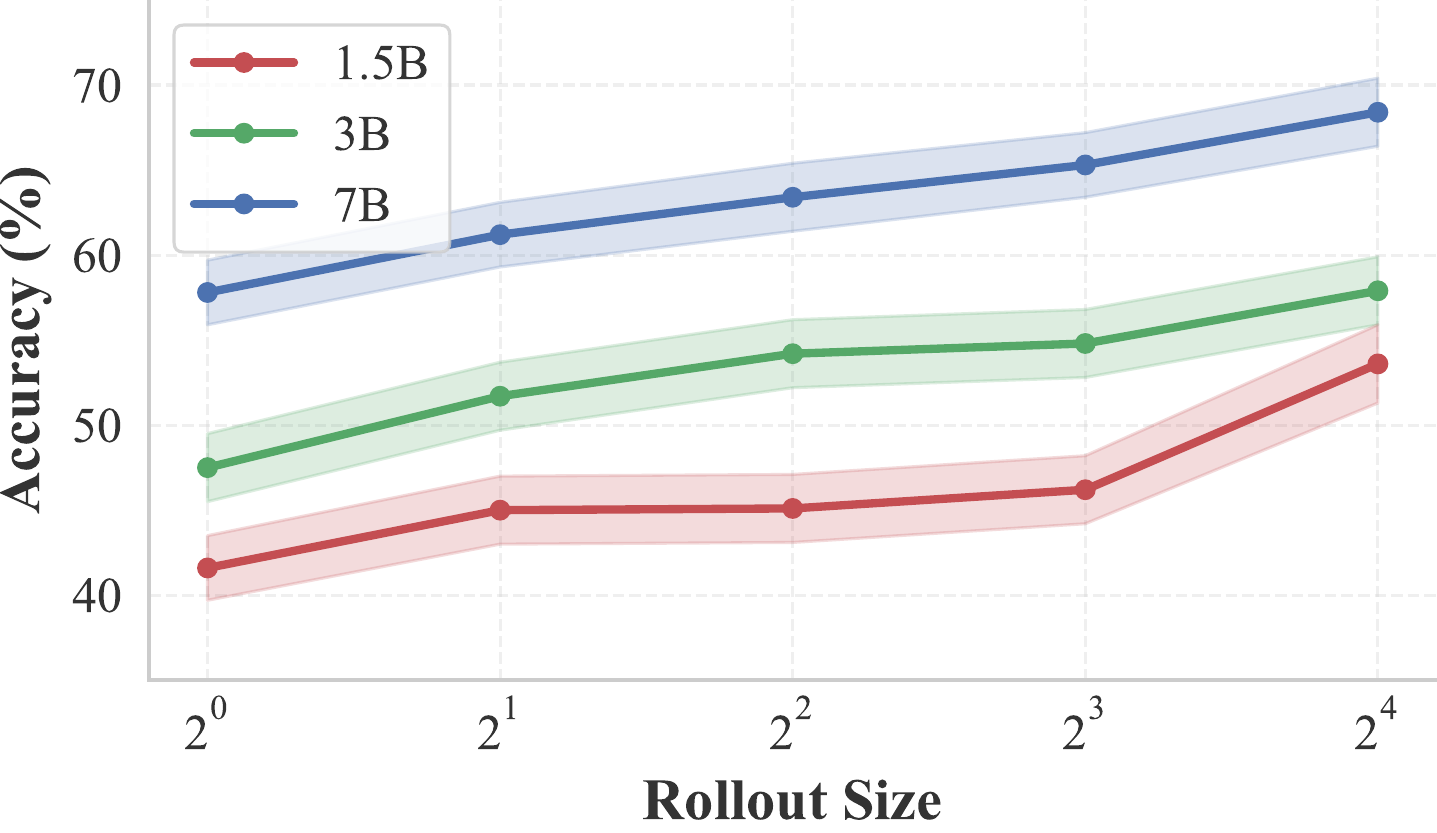}
  \caption{\textbf{Test-time Scaling (TTS) on DeepMath.} Performance improves as the number of rollouts ($N$) increases for all model sizes. See Table~\ref{tab:deepmath_tts_raro} in Appendix~\ref{sec:additional_tables_figures} for detailed data.}
  \label{fig:deepmath_tts}
  \vspace{1ex}
\end{figure}

Another key advantage of RARO is that our learned critic enables natural Test-Time Scaling (TTS) to further improve the policy's performance. 
Specifically, the critic's pairwise comparison setup allows for a \emph{single-elimination tournament} with the critic as the judge (see Algorithm~\ref{alg:tts_tournament} in Appendix~\ref{sec:implementation_details_tts}), enabling further policy improvements with additional rollouts. 
As shown in Figure~\ref{fig:deepmath_tts}, increasing the number of rollouts from 1 to 16 consistently improves performance. 
In particular, with 16 rollouts, RARO achieves $53.6\%$ on the 1.5B model and $57.9\%$ on the 3B model. 
Compared with the RLVR baseline with the same TTS strategy (Table~\ref{tab:deepmath_tts_raro} and Table~\ref{tab:deepmath_tts_rlvr} in the Appendix), RARO achieves a similar rate of improvement.
This result highlights that RARO, when combined with test-time search, can scale effectively, matching the scaling trends of models trained with oracle verifiers.

\subsection{Poetry Writing}

\begin{figure}[t]
        \centering
        \includegraphics[width=\linewidth]{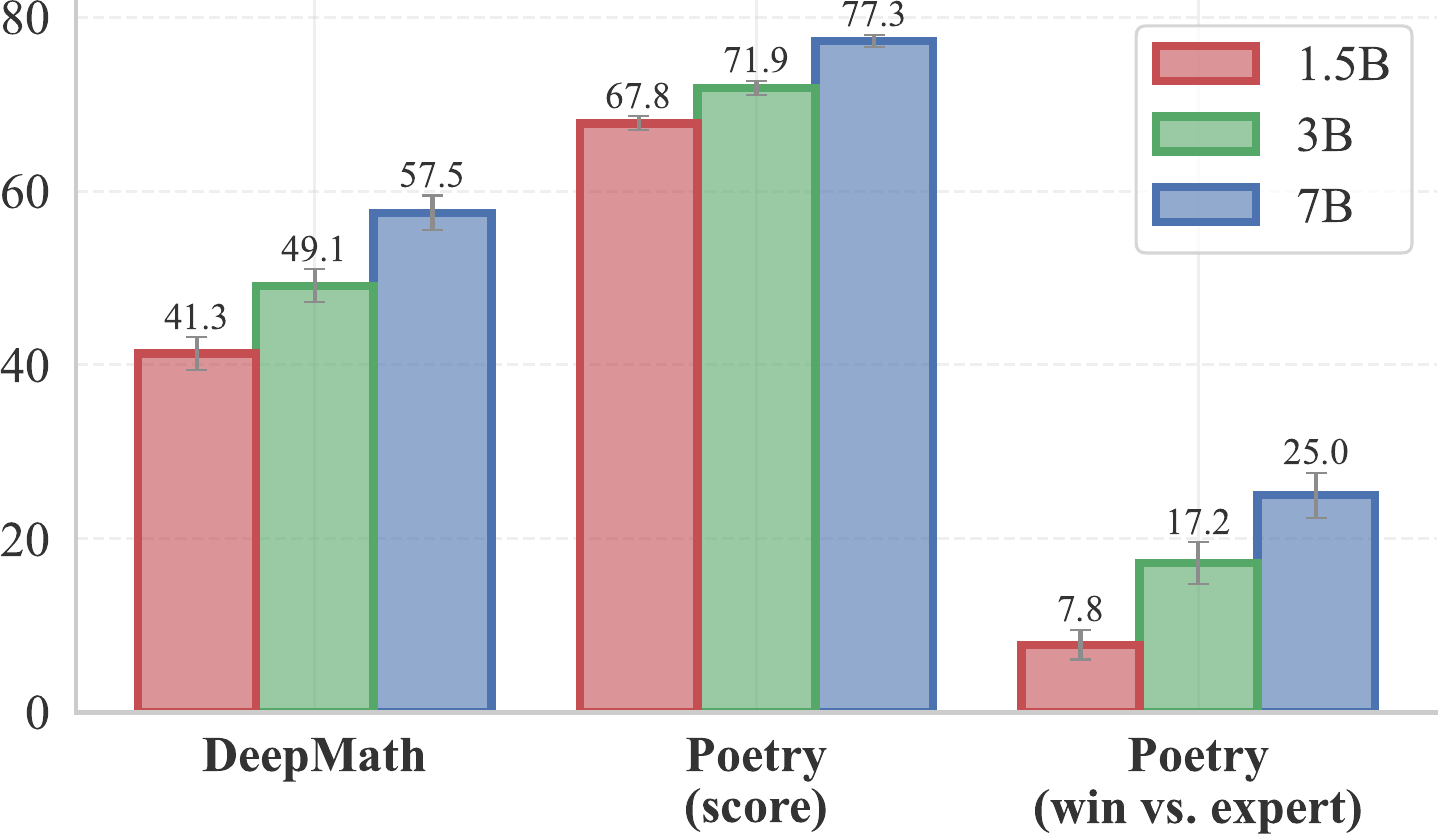}
        \caption{Performance scaling of RARO across model sizes.}
        \label{fig:model_size_scale}
\end{figure}

Finally, we study RARO on Poetry Writing, an open-ended, non-verifiable domain that benefits from specialized reasoning capabilities. Similarly to DeepMath, we study the performance of RARO across a range of model sizes.

\vspace{-1ex}
\paragraph{Surpassing Supervised Baselines.}
Table~\ref{tab:main_results} reveals a substantial performance gap between RARO and baselines. 
Although SFT and Rationalization achieve modest win rates against expert poetry (peaking at $5.9\%$ with the 7B model), RARO reaches $25.0\%$, a nearly 4x improvement.
This advantage is also reflected in the scoring evaluation, where RARO consistently surpasses baselines (e.g., $67.8$ vs. $53.7$ for SFT at 1.5B). 
Notably, RL-Logit, the leading baseline for DeepMath, does not produce competitive results, yielding a near-zero improvement over the base model ($36.4$ vs. $35.0$ at 1.5B) for both the win rate and the scoring evaluation.

\vspace{-1ex}
\paragraph{Scaling with Model Size.}
A key result is that RARO scales with the model size in the creative domain
of poetry, as shown in Figure~\ref{fig:model_size_scale}. 
As we increase the model capacity from 1.5B to 7B, the win rate against expert human poems grows substantially, from $7.8\%$ to $25.0\%$. 
The scoring evaluation also improves from $67.8$ to $77.3$. 
This trend shows that, similarly to verifiable domains, RARO continues to effectively scale with model size in open-ended domains.

\vspace{-1ex}
\paragraph{Emergent Reasoning.}
Qualitatively, RARO induces explicit planning and reasoning behaviors. 
As shown in Figure~\ref{fig:qualitative_examples} in Appendix~\ref{sec:additional_tables_figures}, the model learns to decompose complex problems into structured plans.
Specifically, in poetry writing, it decomposes the prompt into key themes (e.g., \enquote{sensory details}) and stylistic constraints (e.g., \enquote{free verse}, \enquote{unrhymed, natural cadences}) before generating the poem. 
This shows that RARO effectively elicits reasoning that aligns the model's output to the task requirements.

\vspace{-2ex}
\section{Ablation Studies}

\paragraph{Contribution of Each Component}

\begin{table}[t]
    \caption{\textbf{Ablation results on DeepMath 1.5B.} Removing any component leads to performance degradation.}
    \label{tab:ablation_deepmath}
    \centering
    \small

    \begin{tabular}{l c}
        \toprule
        Method & DeepMath 1.5B accuracy ($\%$) $\uparrow$\\
        \midrule
        without critic reasoning      & $35.9 \pm 1.9$ \\
        without relativistic critic   & $36.9 \pm 1.9$ \\
        without tie option            & $38.6 \pm 1.9$ \\
        without replay buffer         & $35.4 \pm 1.8$ \\
        without shared LLM            & $39.4 \pm 1.9$ \\
        \midrule
        \textbf{RARO}             & $\mathbf{41.3 \pm 1.9}$ \\
        \bottomrule
    \end{tabular}
\end{table}

We conduct Leave-One-Out (LOO) ablations on the DeepMath dataset at 1.5B to isolate the contribution of each component in our framework. 
As summarized in Table~\ref{tab:ablation_deepmath}, removing any single component results in a significant performance degradation. 
This uniform drop confirms that all designed mechanisms are essential for the method's overall effectiveness.

Beyond aggregate metrics, we observe distinct failure modes associated with particular missing components, illustrated by the training dynamics in Appendix~\ref{sec:ablation_studies}.

\vspace{-2ex}
\paragraph{Hyperparameter Sensitivity}
\label{app:hparam-sensitivity}

We test whether RARO depends on a narrow choice of its main task-dependent coefficients. 
As shown in Table~\ref{tab:hparam-sensitivity}, validation performance remains stable across reasonable variations of the tie rewards on DeepMath and the policy-loss weight on poetry. 
Together with the above, these results suggest that once the replay buffer, tie option, critic reasoning, and relativistic comparison are enabled, RARO does not rely on a single finely tuned coefficient setting.

\begin{table}[h]
\centering
\small
\caption{
Hyperparameter sensitivity for the main task-dependent RARO coefficients.
}
\label{tab:hparam-sensitivity}
\begin{tabular}{llcc}
\toprule
Task & Coefficient & Value & Validation \\
\midrule
\multirow{3}{*}{DeepMath 1.5B}
& \multirow{3}{*}{$\tau_{\mathrm{crit}}$}
& $0.52$ & $41.1$ acc. \\
& & $0.55$ & $42.0$ acc. \\
& & $0.60$ & $40.7$ acc. \\
\midrule
\multirow{3}{*}{DeepMath 1.5B}
& \multirow{3}{*}{$\tau_{\mathrm{pol}}$}
& $0.20$ & $40.8$ acc. \\
& & $0.40$ & $40.2$ acc. \\
& & $0.60$ & $42.0$ acc. \\
\midrule
\multirow{3}{*}{Poetry 1.5B}
& \multirow{3}{*}{$\lambda_{\mathrm{pol}}$}
& $4/5$ & $64.0$ reward \\
& & $2/3$ & $68.8$ reward \\
& & $1/3$ & $66.2$ reward \\
\bottomrule
\end{tabular}
\end{table}
\section{Conclusion \& Future Work}

We introduce RARO (Relativistic Adversarial Reasoning Optimization), a novel approach to training reasoning LLMs using only expert demonstrations, thereby bypassing the need for task-specific verifiers or expensive preference annotations. By formulating the problem as Inverse Reinforcement Learning and incorporating a relativistic critic setup, we obtain a principled and stable adversarial training algorithm that yields strong reasoning capabilities.

Our experiments demonstrate the effectiveness of RARO in multiple setups.
On the controlled Countdown task, it not only outperforms verifier-free baselines but also nearly matches the performance of RLVR. 
In the general math domain, it exhibits similar scalability trends to RLVR while outperforming baselines without verification. 
Lastly, on the open-ended Poetry Writing task, it successfully elicits specialized reasoning capabilities and significantly surpasses all baselines.
Together, these findings suggest that RARO is a promising and practical approach for training reasoning models without relying on explicit verifiers.

Future work includes extending the framework to more generalized adversarial setups that stabilize training across diverse domains, improving sample efficiency, scaling the method to larger, state-of-the-art model sizes, and developing an alternative critic setup that enables better reward interpretability. See Appendix \ref{sec:future_work} for more details.

\section*{Impact Statement}
This work advances methods for training reasoning-capable language models using expert demonstrations without task-specific verifiers. 
The primary expected benefits are improved sample efficiency and broader applicability of reasoning learning to domains where automatic verification is not available, which may reduce reliance on costly human preference collection. 
Potential risks mirror those of more capable language models in general, none of which we feel must be specifically highlighted here. 
Our method does not introduce new data sources or deployment mechanisms, but it may lower the barrier to training stronger models in non-verifiable settings. 
We recommend standard mitigations for responsible release and deployment, including careful dataset curation, safety evaluations prior to deployment, and usage policies aligned with the application domain.

\bibliography{bibliography}
\bibliographystyle{icml2026}

\appendix
\onecolumn
\section{Derivations}\label{apn:derivations}

\subsection{Derivation of Closed-Form Optimal Policy}\label{apn:derivations:closed-form-policy}

\begin{proposition}
    Consider the KL-regularized reward-maximization objective:
    \[
    \theta^\star(\phi)
    =\arg\max_{\theta}\;
    \mathbb{E}_{(q,a)\sim \hat{p}_D}
    \Big[
      r_\phi(a,q)
      - \beta\, D_{\mathrm{KL}}\!\big(\pi_\theta(\cdot \mid q)\,\|\,\pi_{\mathrm{ref}}(\cdot \mid q)\big)
    \Big].
    \]
    The optimal policy has the following closed-form solution:
    \[
    \pi_{\theta^\star(\phi)}(a \mid q) 
    = \frac{1}{Z_{\theta^\star(\phi)}(q)} \, 
    \pi_{\mathrm{ref}}(a \mid q) \,
    \exp\!\Bigg\{\frac{1}{\beta} \, r_\phi(a,q)\Bigg\},
    \]
    where \(Z_{\theta^\star(\phi)}(q)\) is the partition function ensuring normalization.
\end{proposition}

\begin{proof}
We derive the closed-form solution for the KL-regularized reward maximization objective. Consider the objective function for a single question $q$:
\begin{equation}
    \mathcal{J}(\pi) = \mathbb{E}_{a \sim \pi(\cdot|q)} \left[ r_\phi(a,q) \right] - \beta D_{\mathrm{KL}}\big(\pi(\cdot \mid q)\,\|\,\pi_{\mathrm{ref}}(\cdot \mid q)\big).
\end{equation}
Expanding the KL divergence term:
\begin{align*}
    D_{\mathrm{KL}}(\pi \| \pi_{\mathrm{ref}}) &= \mathbb{E}_{a \sim \pi(\cdot|q)} \left[ \log \frac{\pi(a \mid q)}{\pi_{\mathrm{ref}}(a \mid q)} \right] \\
    &= \mathbb{E}_{a \sim \pi(\cdot|q)} \left[ \log \pi(a \mid q) - \log \pi_{\mathrm{ref}}(a \mid q) \right].
\end{align*}
Substituting this back into the objective:
\begin{align*}
    \mathcal{J}(\pi) &= \mathbb{E}_{a \sim \pi(\cdot|q)} \left[ r_\phi(a,q) - \beta \log \pi(a \mid q) + \beta \log \pi_{\mathrm{ref}}(a \mid q) \right] \\
    &= \beta \, \mathbb{E}_{a \sim \pi(\cdot|q)} \left[ \frac{1}{\beta} r_\phi(a,q) + \log \pi_{\mathrm{ref}}(a \mid q) - \log \pi(a \mid q) \right] \\
    &= -\beta \, \mathbb{E}_{a \sim \pi(\cdot|q)} \left[ \log \pi(a \mid q) - \left( \log \pi_{\mathrm{ref}}(a \mid q) + \frac{1}{\beta} r_\phi(a,q) \right) \right].
\end{align*}
Let us define the normalized Gibbs distribution:
\begin{equation}
    \pi^*(a \mid q) = \frac{1}{Z(q)} \pi_{\mathrm{ref}}(a \mid q) \exp\left( \frac{1}{\beta} r_\phi(a,q) \right),
\end{equation}
where $Z(q) = \int \pi_{\mathrm{ref}}(a' \mid q) \exp\left( \frac{1}{\beta} r_\phi(a',q) \right) da'$ is the partition function.
Taking the logarithm of $\pi^*$:
\begin{equation}
    \log \pi^*(a \mid q) = \log \pi_{\mathrm{ref}}(a \mid q) + \frac{1}{\beta} r_\phi(a,q) - \log Z(q).
\end{equation}
Substituting $\log \pi_{\mathrm{ref}}(a \mid q) + \frac{1}{\beta} r_\phi(a,q) = \log \pi^*(a \mid q) + \log Z(q)$ into the objective:
\begin{align*}
    \mathcal{J}(\pi) &= -\beta \, \mathbb{E}_{a \sim \pi(\cdot|q)} \left[ \log \pi(a \mid q) - \big( \log \pi^*(a \mid q) + \log Z(q) \big) \right] \\
    &= -\beta \left( \mathbb{E}_{a \sim \pi(\cdot|q)} \left[ \log \frac{\pi(a \mid q)}{\pi^*(a \mid q)} \right] - \log Z(q) \right) \\
    &= -\beta D_{\mathrm{KL}}(\pi \| \pi^*) + \beta \log Z(q).
\end{align*}
Since $\beta > 0$ and $\log Z(q)$ does not depend on $\pi$, maximizing $\mathcal{J}(\pi)$ is equivalent to minimizing the KL divergence $D_{\mathrm{KL}}(\pi \| \pi^*)$. By Gibbs' inequality, $D_{\mathrm{KL}}(\pi \| \pi^*) \ge 0$, with equality if and only if $\pi = \pi^*$ almost everywhere.
Thus, the optimal policy is given by:
\begin{equation}
    \pi_{\theta^\star(\phi)}(a \mid q) = \pi^*(a \mid q) = \frac{1}{Z(q)} \pi_{\mathrm{ref}}(a \mid q) \exp\left( \frac{1}{\beta} r_\phi(a,q) \right).
\end{equation}
\end{proof}

\subsection{Proof of Reward Gradient}\label{apn:derivations:reward-grad}

\begin{proposition}
    Using the closed-form policy, the gradient of the maximum likelihood objective $\mathcal{L}(\phi) = \mathbb{E}_{(q,a)\sim \hat{p}_D}\!\left[\log \pi_{\theta^\star(\phi)}(a \mid q)\right]$ with respect to $\phi$ is:
    \[
    \nabla_\phi \mathcal{L}(\phi)
    = \frac{1}{\beta}\Big(
    \mathbb{E}_{(q,a)\sim \hat{p}_D}\big[\nabla_\phi r_\phi(a,q)\big]
    -\mathbb{E}_{q\sim \hat{p}_q}\,\mathbb{E}_{a'\sim \pi_{\theta^\star(\phi)}(\cdot\mid q)}\big[\nabla_\phi r_\phi(a',q)\big]
    \Big).
    \]
\end{proposition}

\begin{proof}
We aim to derive the gradient of the data log-likelihood objective with respect to the reward parameters $\phi$. Recall the objective:
\begin{equation}
    \mathcal{L}(\phi) = \mathbb{E}_{(q,a)\sim \hat{p}_D}[\log \pi_{\theta^\star(\phi)}(a \mid q)].
\end{equation}
The optimal policy $\pi_{\theta^\star(\phi)}$ takes the closed-form solution:
\begin{equation}
    \pi_{\theta^\star(\phi)}(a \mid q) = \frac{1}{Z_{\theta^\star(\phi)}(q)} \pi_{\mathrm{ref}}(a \mid q) \exp\left(\frac{1}{\beta} r_\phi(a,q)\right),
\end{equation}
where $Z_{\theta^\star(\phi)}(q) = \int \pi_{\mathrm{ref}}(a' \mid q) \exp\left(\frac{1}{\beta} r_\phi(a',q)\right) da'$ is the partition function.

Substituting the policy expression into the log-likelihood:
\begin{align*}
    \log \pi_{\theta^\star(\phi)}(a \mid q) &= \log \left( \frac{\pi_{\mathrm{ref}}(a \mid q) \exp\left(\frac{1}{\beta} r_\phi(a,q)\right)}{Z_{\theta^\star(\phi)}(q)} \right) \\
    &= \log \pi_{\mathrm{ref}}(a \mid q) + \frac{1}{\beta} r_\phi(a,q) - \log Z_{\theta^\star(\phi)}(q).
\end{align*}
Since $\pi_{\mathrm{ref}}$ does not depend on $\phi$, the gradient is:
\begin{align*}
    \nabla_\phi \log \pi_{\theta^\star(\phi)}(a \mid q) &= \nabla_\phi \left( \frac{1}{\beta} r_\phi(a,q) - \log Z_{\theta^\star(\phi)}(q) \right) \\
    &= \frac{1}{\beta} \nabla_\phi r_\phi(a,q) - \frac{\nabla_\phi Z_{\theta^\star(\phi)}(q)}{Z_{\theta^\star(\phi)}(q)}.
\end{align*}
We now compute the gradient of the partition function $Z_{\theta^\star(\phi)}(q)$ using the Leibniz integral rule (interchanging gradient and integral):
\begin{align*}
    \nabla_\phi Z_{\theta^\star(\phi)}(q) &= \nabla_\phi \int \pi_{\mathrm{ref}}(a' \mid q) \exp\left(\frac{1}{\beta} r_\phi(a',q)\right) da' \\
    &= \int \pi_{\mathrm{ref}}(a' \mid q) \nabla_\phi \exp\left(\frac{1}{\beta} r_\phi(a',q)\right) da' \\
    &= \int \pi_{\mathrm{ref}}(a' \mid q) \exp\left(\frac{1}{\beta} r_\phi(a',q)\right) \left( \frac{1}{\beta} \nabla_\phi r_\phi(a',q) \right) da'.
\end{align*}
Substituting this back into the gradient term for $\log Z_{\theta^\star(\phi)}(q)$:
\begin{align*}
    \frac{\nabla_\phi Z_{\theta^\star(\phi)}(q)}{Z_{\theta^\star(\phi)}(q)} &= \int \frac{\pi_{\mathrm{ref}}(a' \mid q) \exp\left(\frac{1}{\beta} r_\phi(a',q)\right)}{Z_{\theta^\star(\phi)}(q)} \left( \frac{1}{\beta} \nabla_\phi r_\phi(a',q) \right) da' \\
    &= \int \pi_{\theta^\star(\phi)}(a' \mid q) \left( \frac{1}{\beta} \nabla_\phi r_\phi(a',q) \right) da' \\
    &= \mathbb{E}_{a'\sim \pi_{\theta^\star(\phi)}(\cdot\mid q)}\left[ \frac{1}{\beta} \nabla_\phi r_\phi(a',q) \right].
\end{align*}
Finally, averaging over the dataset $(q,a) \sim \hat{p}_D$:
\begin{align*}
    \nabla_\phi \mathcal{L}(\phi) &= \mathbb{E}_{(q,a)\sim \hat{p}_D} \left[ \frac{1}{\beta} \nabla_\phi r_\phi(a,q) - \mathbb{E}_{a'\sim \pi_{\theta^\star(\phi)}(\cdot\mid q)}\left[ \frac{1}{\beta} \nabla_\phi r_\phi(a',q) \right] \right] \\
    &= \frac{1}{\beta} \left( \mathbb{E}_{(q,a)\sim \hat{p}_D}[\nabla_\phi r_\phi(a,q)] - \mathbb{E}_{q\sim \hat{p}_q} \mathbb{E}_{a'\sim \pi_{\theta^\star(\phi)}(\cdot\mid q)}[\nabla_\phi r_\phi(a',q)] \right).
\end{align*}
This completes the derivation.
\end{proof}

\subsection{Derivation of Reasoning Reward Gradient}\label{apn:derivations:reasoning-reward}

\begin{proposition}
    Let the reward be parameterized by a critic $c_{\phi}(\ell \mid a,q)$ with labels $\ell \in \{\mathrm{expert}, \mathrm{policy}\}$ as $r_\phi(a,q) = c_{\phi}(\ell=\mathrm{expert} \mid a,q)$. The gradient of the loss with respect to critic parameters \(\phi\) is:
    \[
    \nabla_{\phi}L = 
    \frac{1}{\beta} \, \mathbb{E}_{q \sim \hat{p}_q(\cdot)} \!
    \Bigg[
      \mathbb{E}_{a \sim \hat{p}_{a\mid q}(\cdot \mid q) \,\cup\, \pi_\theta(a \mid q)}
      \Big[
        \mathbb{E}_{\ell \sim c_\phi(\cdot \mid a,q)}
        \Big[
          R(\ell,a,q) \, \nabla_{\phi} \log c_\phi(\ell \mid a,q)
        \Big]
      \Big]
    \Bigg],
    \]
    where
    \[
    R(\ell,a,q) = \mathbbm{1}_{\ell \text{ is correct}}
    \]
\end{proposition}

\begin{proof}
In this section, we derive the specific form of the reward gradient when the reward is parameterized by a critic LLM $c_\phi$.
Recall from Eq. (\ref{apn:derivations:reward-grad}) that the gradient of the loss is:
\[
\nabla_\phi \mathcal{L}(\phi) = \frac{1}{\beta} \left( \mathbb{E}_{(q,a)\sim \hat{p}_D}[\nabla_\phi r_\phi(a,q)] - \mathbb{E}_{q\sim \hat{p}_q} \mathbb{E}_{a'\sim \pi_{\theta}(\cdot\mid q)}[\nabla_\phi r_\phi(a',q)] \right),
\]
where we have approximated the optimal policy $\pi_{\theta^\star(\phi)}$ with the current policy $\pi_\theta$.

We parameterize the reward using the probability of the answer being expert-generated, as predicted by a binary classifier (critic) $c_\phi(\ell \mid a, q)$ where $\ell \in \{\mathrm{expert}, \mathrm{policy}\}$:
\[
r_\phi(a,q) = c_{\phi}(\ell=\mathrm{expert} \mid a,q).
\]
Let $p_E = c_\phi(\ell=\mathrm{expert} \mid a, q)$ and $p_P = c_\phi(\ell=\mathrm{policy} \mid a, q) = 1 - p_E$.
The gradient of the reward with respect to $\phi$ is:
\[
\nabla_\phi r_\phi(a,q) = \nabla_\phi p_E.
\]
We can express this gradient using the REINFORCE trick (log-derivative trick) over the binary outcome $\ell$. Consider the quantity:
\begin{align*}
    \mathbb{E}_{\ell \sim c_\phi(\cdot \mid a,q)} [ \mathbbm{1}_{\ell=\mathrm{expert}} \nabla_\phi \log c_\phi(\ell \mid a,q) ]
    &= \mathbbm{1}_{\mathrm{expert}=\mathrm{expert}} p_E \nabla_\phi \log p_E + \mathbbm{1}_{\mathrm{policy}=\mathrm{expert}} p_P \nabla_\phi \log p_P \\
    &= 1 \cdot \nabla_\phi p_E + 0 \\
    &= \nabla_\phi p_E.
\end{align*}
Thus, we have the identity:
\[
\nabla_\phi r_\phi(a,q) = \mathbb{E}_{\ell \sim c_\phi(\cdot \mid a,q)} \Big[ \mathbbm{1}_{\ell=\mathrm{expert}} \nabla_\phi \log c_\phi(\ell \mid a,q) \Big].
\]

Substituting this identity back into the loss gradient expression:

1. \textbf{Expert Term} ($(q,a) \sim \hat{p}_D$):
The answers come from the expert distribution, so the correct label is $\mathrm{expert}$.
\begin{align*}
\mathbb{E}_{a \sim \hat{p}_{a\mid q}} [\nabla_\phi r_\phi(a,q)] 
&= \mathbb{E}_{a \sim \hat{p}_{a\mid q}} \Big[ \mathbb{E}_{\ell \sim c_\phi} \Big[ \mathbbm{1}_{\ell=\mathrm{expert}} \nabla_\phi \log c_\phi(\ell \mid a,q) \Big] \Big].
\end{align*}
This corresponds to a reward signal of $+1$ when $\ell=\mathrm{expert}$ (correct) and $0$ when $\ell=\mathrm{policy}$ (incorrect).

2. \textbf{Policy Term} ($a \sim \pi_\theta$):
Note the negative sign in the original gradient formula.
\begin{align*}
- \mathbb{E}_{a \sim \pi_\theta} [\nabla_\phi r_\phi(a,q)] 
&= - \mathbb{E}_{a \sim \pi_\theta} [\nabla_\phi p_E] \\
&= \mathbb{E}_{a \sim \pi_\theta} [\nabla_\phi p_P] \quad (\text{since } \nabla_\phi p_E + \nabla_\phi p_P = 0) \\
&= \mathbb{E}_{a \sim \pi_\theta} \Big[ \mathbb{E}_{\ell \sim c_\phi} \Big[ \mathbbm{1}_{\ell=\mathrm{policy}} \nabla_\phi \log c_\phi(\ell \mid a,q) \Big] \Big].
\end{align*}
Here, the answers come from the policy, so the correct label is $\mathrm{policy}$. This corresponds to a reward signal of $+1$ when $\ell=\mathrm{policy}$ (correct) and $0$ when $\ell=\mathrm{expert}$ (incorrect).

Combining both terms and grouping the expectations results in the final expression:
\[
\nabla_{\phi}L = \frac{1}{\beta} \, \mathbb{E}_{q \sim \hat{p}_q} \Bigg[
  \mathbb{E}_{a \sim \hat{p}_{a\mid q} \cup \pi_\theta}
  \Big[
    \mathbb{E}_{\ell \sim c_\phi}
    \Big[
      R(\ell,a,q) \, \nabla_{\phi} \log c_\phi(\ell \mid a,q)
    \Big]
  \Big]
\Bigg],
\]
where the reward $R(\ell,a,q)$ aggregates the signs from both cases:
    \[
    R(\ell,a,q) = \mathbbm{1}_{\ell \text{ is correct}}
    \]
\end{proof}
\section{Future Work} \label{sec:future_work}

\begin{figure}[ht]
    \centering
    \includegraphics[width=0.42\textwidth]{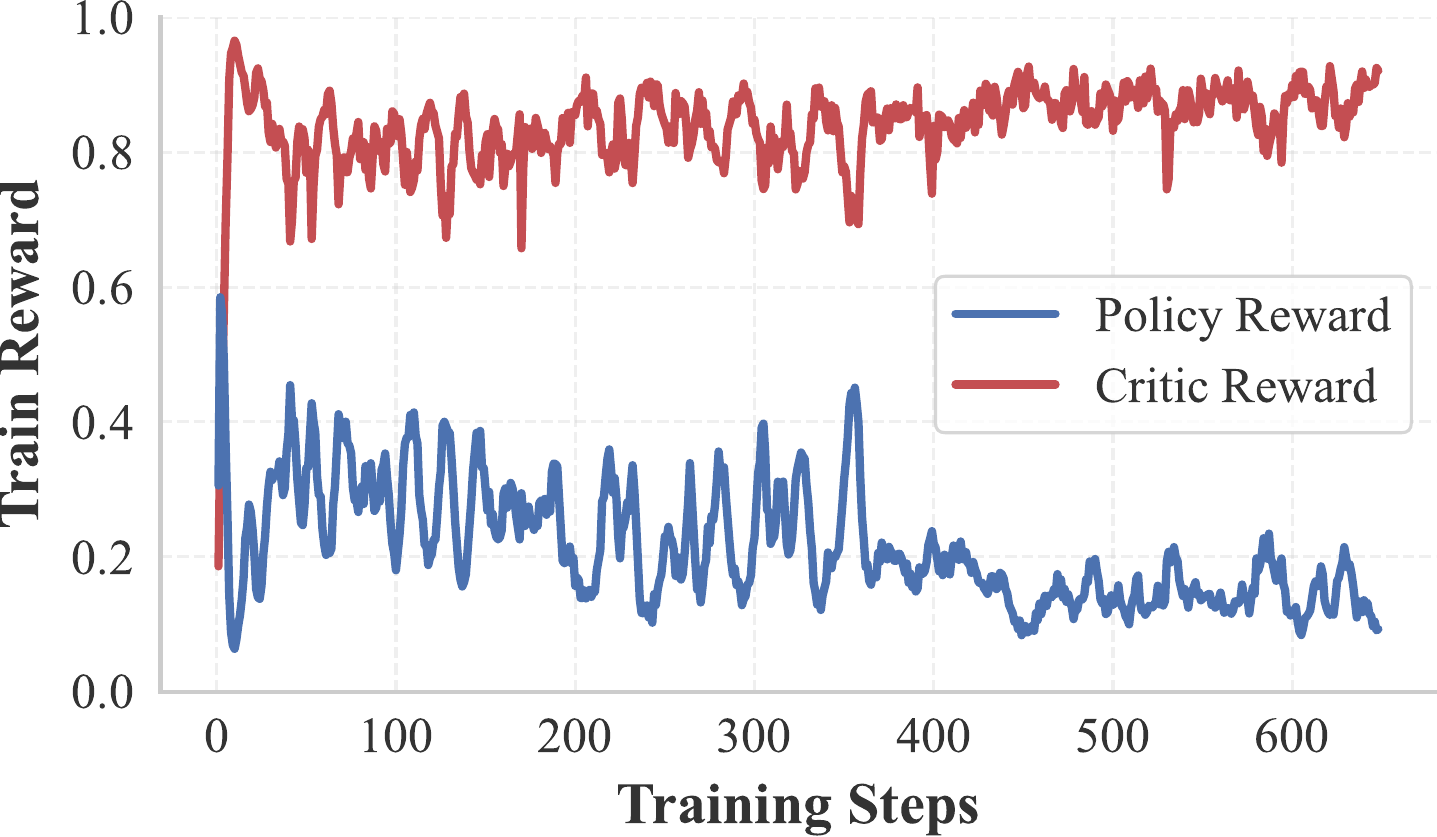}
    \hfill
    \includegraphics[width=0.42\textwidth]{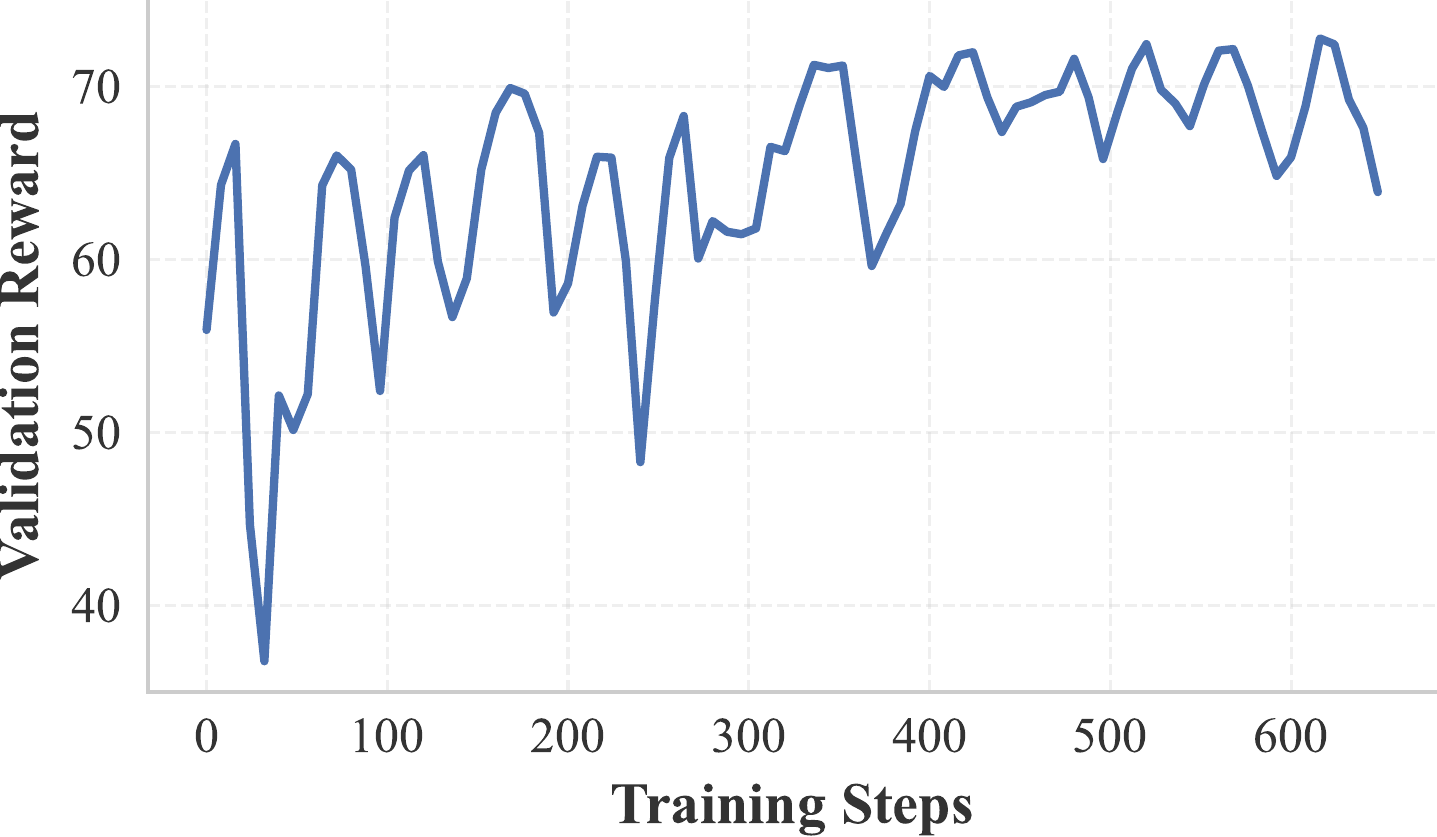}
    \caption{\textbf{Poetry Writing (7B) Training Dynamics.} During training, the policy and critic rewards oscillate on the Poetry Writing task (left). The validation reward similarly oscillates despite an overall upward trend (right).}
    \label{fig:poetry_rewards}
\end{figure}

\begin{figure}[ht]
    \centering
    \begin{minipage}{0.48\textwidth}
        \centering
        \includegraphics[width=0.8\linewidth]{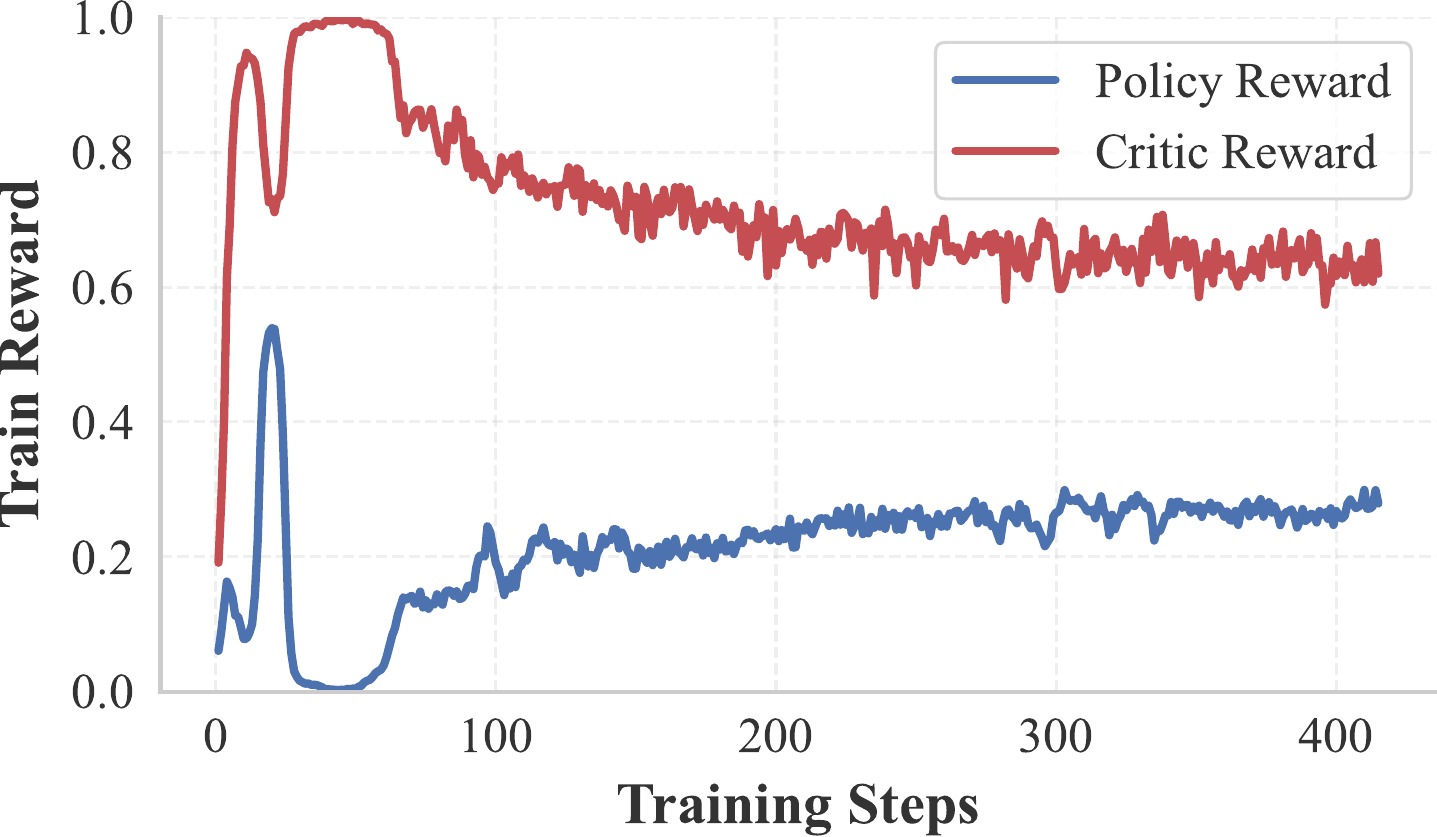}
    \end{minipage}
    \hfill
    \begin{minipage}{0.48\textwidth}
        \centering
        \includegraphics[width=0.8\linewidth]{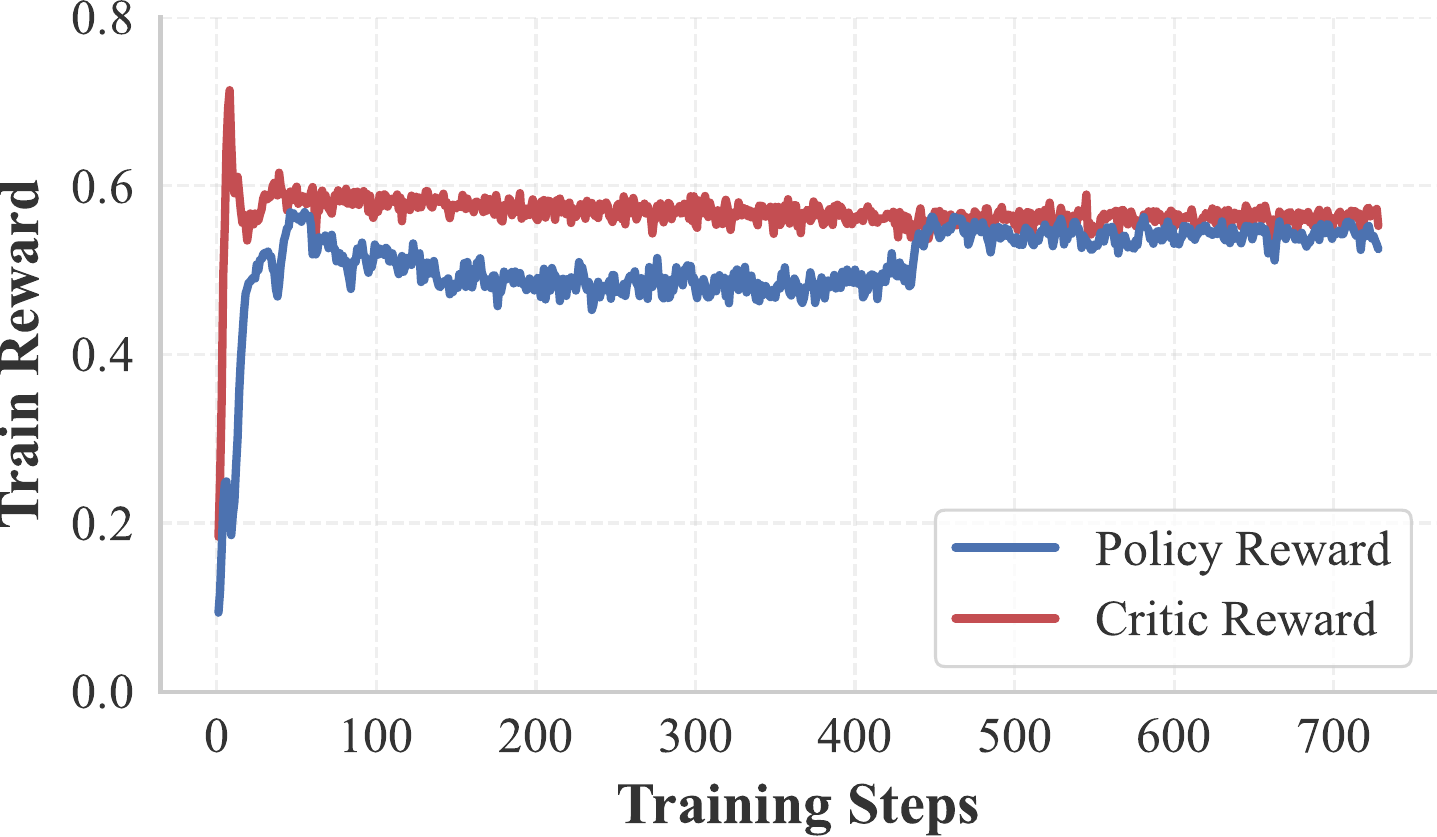}
    \end{minipage}
    \caption{\textbf{Countdown and DeepMath (1.5B) Training Dynamics.} Stable policy and critic rewards during training for Countdown and DeepMath.}
    \label{fig:training_dynamics_1.5b}
\end{figure}

\paragraph{Stability in Long-form Generation.} While RARO exhibits stable training dynamics on verifiable tasks (see Figure~\ref{fig:training_dynamics_1.5b}), we observed some instability in long-form creative tasks. As shown in Figure~\ref{fig:poetry_rewards}, during training, the policy and critic rewards could oscillate on the Poetry Writing task. Additionally, the validation reward similarly oscillates despite an overall upward trend. This is reminiscent of the instability observed in adversarial training for generative models, where powerful discriminators can overfit to transient artifacts and induce non-stationary learning dynamics for the generator ~\citep{karras2020traininggenerativeadversarialnetworks}. Future work will focus on developing techniques to stabilize this adversarial game in subjective domains. It will also be important to understand when such oscillations reflect true ambiguity in the task (e.g., multiple equally valid poetic styles) versus undesirable instability that harms downstream user experience.

\begin{figure}[t]
    \centering
    \includegraphics[width=0.42\textwidth]{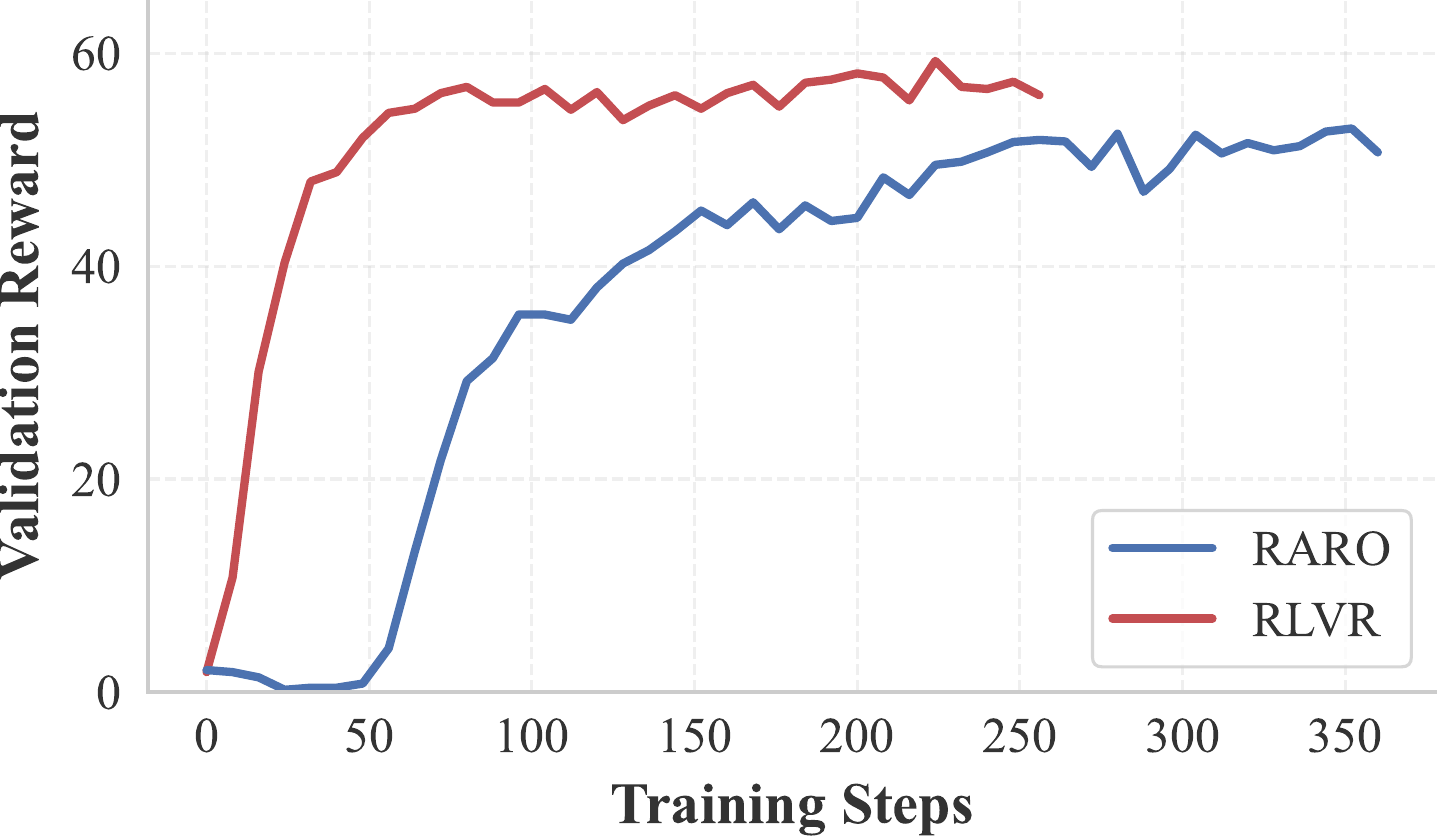}
    \caption{\textbf{Sample Efficiency Comparison.}
    Under the same hyperparameters, our method is less sample-efficient than RLVR on Countdown.}
    \label{fig:inefficiency}
\end{figure}

\paragraph{Sample Efficiency.} While RARO achieves strong final performance, it can be less sample-efficient than RLVR when applied to verifiable tasks. As shown in Figure~\ref{fig:inefficiency}, under identical hyperparameters on the Countdown task, RARO requires more training iterations to reach performance levels comparable to RLVR. This inefficiency stems from the added complexity of jointly training a policy and critic in an adversarial game, where the critic must first learn to discriminate between policy and expert answers before providing a useful training signal. In contrast, RLVR benefits from immediate, oracle feedback. While this gap is unavoidable without access to a ground-truth verifier, future work could explore techniques to accelerate convergence, such as curriculum learning and critic pretraining. A complementary direction is to theoretically characterize the sample complexity of our relativistic adversarial objective and identify conditions under which we can provably bound the sample complexity of RARO.

\paragraph{Reward Interpretability.} One motivation for our critic design is to produce natural-language feedback that resembles human-written explanations. However, even when the critic outputs detailed justifications, it remains challenging to extract a compact, stable, and explicitly human-interpretable rubric that governs its behavior. In practice, the critic's preferences may be entangled across many latent factors, and at different training steps, the critic may prefer vastly different answers. Making the critic truly interpretable therefore remains an open problem at the intersection of IRL, interpretability, and value learning: promising directions include probing critic representations for concept-like features and distilling the critic into simpler rubric models.

\paragraph{Scaling Reasoning.} We aim to scale RARO to larger base models beyond 7B parameters and beyond reasoning budget of 2048 tokens. Our results already indicate that increasing the reasoning budget---via longer chains of thought at train time and test time---can yield substantial gains. Thus, we are interested in exploring how scaling our method both in model size and reasoning budget can lead to new emerging reasoning capabilities. Another important direction is to finetune models that already exhibit strong reasoning capabilities on new tasks using RARO, so that they can rapidly adapt their reasoning strategies without requiring task-specific verifiers or human preference labels. 

\paragraph{Broadening Non-verifiable Domains.} Finally, we plan to apply our approach to a wider range of open-ended domains, such as front-end software development and long-form scientific writing, where expert demonstrations are plentiful online but reliable verifiers are absent. If successful, our approach could enable a new wave of practical LLM applications in these domains, unlocking capabilities where training signals were previously scarce or unreliable. This would allow for the deployment of robust reasoning systems in complex, real-world environments without the need for expensive or impossible-to-design verifiers.
\section{Implementation Details} \label{sec:implementation_details}

\subsection{Stable \& Efficient Learning} \label{sec:implementation_details_stable_learning}

Here we describe the specific techniques that enable stable and efficient learning in RARO.

\paragraph{Shared LLM for Critic and Policy.}
While Section \ref{sec:method:relativistic-critic} provides a practical procedure for alternating updates of the policy (\(\theta\)) and the critic model (\(\phi\)), it requires training \emph{two} reasoning LLMs and thus incurs long, token-intensive rollouts for both. To reduce memory usage and potentially promote generalization via shared representations, we ultimately use the same underlying LLM to role-play as both the critic and the policy. Our ablations (see Appendix \ref{sec:ablation_studies}) empirically support that using a shared LLM for the critic and the policy improves performance.

\paragraph{Data Mixing.}
In addition, by sharing the same underlying LLM, we can substantially simplify the concrete algorithm by \emph{mixing} both the critic and policy rollouts in the same batch to compute advantage and loss. This allows us to remove the need for alternating updates between the critic and the policy and instead perform all updates in a single batch. Furthermore, this setup allows us to easily control the \enquote{strength} of the policy and the critic by adjusting the weight of the critic and policy loss in the combined objective.

\paragraph{Catastrophic Forgetting \& Replay Buffer.}
In GAN training~\citep{gan}, the discriminator often suffers from catastrophic forgetting as the generator \enquote{cycles} among modes to fool it~\citep{gan-catastrophic-forgetting-mode-collapse,gan-continual-learning}. We observe a similar problem in our setting, where policy learns to cycle through a fixed set of strategies to \enquote{hack} the critic reward (see Appendix \ref{sec:ablation_studies}). To mitigate this, we construct half of the critic prompts from a replay buffer of all past policy rollouts, while the other half are sampled from the current batch of policy rollouts, ensuring the critic is continually trained on every mode of \enquote{attack} discovered by the policy. 

In Algorithm~\ref{alg:unified}, the expectation over policy answers is approximated by fresh on-policy rollouts from the current $\pi_\theta$; the replay buffer $\mathcal{R}$ is used only to construct the $(\mathrm{expert}, \mathrm{policy})$ input pairs presented to the critic, not to reuse stale critic outputs or stale policy log-probabilities.

\paragraph{GRPO \& Optimizations.}
Finally, we address several practical issues when implementing the concrete algorithm. When querying the critic to reward policy rollouts, occasional formatting or networking failures produce invalid rewards; we exclude the affected rollouts from the loss by masking them during backpropagation. Following DAPO~\citep{yu2025dapoopensourcellmreinforcement}, we also apply over-length filtering: any policy or critic rollout that exceeds a specified token-length threshold is excluded from the objective computation. Finally, inspired by Dr.~GRPO~\citep{understanding-r1-dr-grpo}, we remove advantage normalization and response-length normalization, which we found to introduce bias in our setting.

\subsection{Datasets} \label{sec:dataset_details}

\paragraph{Countdown.}
We use a 24-style variant of the Countdown arithmetic puzzle, where the goal is to combine four integers using basic arithmetic operations to obtain the target value 24. Instances are synthetically generated via an exhaustive search over all possible combination of operands from $[1, 50]$ and operations from $\{+, -, \times, \div\}$. The instances are then annotated with expert demonstrations by GPT-5~\citep{openai2025gpt5}, discarding instances that GPT-5 cannot solve. The resulting dataset contains 131k total problems, from which we reserve 1024 tasks as a held-out test set. For this task, the final answer is exactly verifiable via a straightforward expression computation, while the underlying search over expressions is substantially more complex.

\paragraph{DeepMath.}
To evaluate our method on general math reasoning domain, we use the DeepMath dataset~\citep{he2025deepmath103klargescalechallengingdecontaminated}, which consists of approximately 103k diverse and high-quality math problems with well-defined ground-truth answers. We utilize the full DeepMath-103K dataset for training and hold out 635 decontaminated problems as a test set. While the dataset provides example reasoning traces beyond ground-truth answers, we discard them in all of our baselines for fairness as our method is not designed to leverage them.

\paragraph{Poetry Writing.}
We construct our poetry writing task from a pre-collected corpus \footnote{\href{https://huggingface.co/datasets/jnb666/poems}{jnb666/poems}} of roughly 40k English-language poems sourced from Poetry Foundation \footnote{\href{https://www.poetryfoundation.org}{Poetry Foundation}}. For each poem, we automatically generate a short human-style prompt using GPT-5 and treat the original poem as the expert demonstration. Out of the 40k poems, we reserve 256 poems at random as our test set. Since poetry writing does not admit an objective verifier, we evaluate RARO and baselines using GPT-5. Specifically, we set up two evaluation metrics: \emph{scalar score} and \emph{win-rate}. The scalar score is measured by prompting GPT-5 to score the poem on a scale of 1-7 then normalized to 0-100, considering both prompt adherence and literary quality. The win-rate is measured by supplying GPT-5 with both the policy and expert poems and prompting it to determine which poem has higher overall quality.

\subsection{Implementation Stack} \label{sec:implementation_details_stack}

\paragraph{Supervised Methods (SFT and Rationalization).}
We train the SFT and Rationalization baselines using Together AI's managed fine-tuning service.\footnote{\url{https://api.together.ai}} While we monitor the validation loss during training, we ultimately select the checkpoint for evaluation based on the best validation reward. 

\paragraph{Iterative Direct Preference Optimization (DPO).}
Our iterative DPO baselines are implemented using the \texttt{trl}~\citep{vonwerra2020trl} library with PyTorch FSDP2 enabled to support efficient distributed training at all model scales. For evaluation, we similarly select the checkpoint that maximizes the validation reward. We repeat this process for 3 rounds.

\paragraph{RL-based methods (RL-Logit, RLVR, and RARO).}
All RL-Based Methods---RL-Logit, RLVR, and RARO---are implemented on top of the \texttt{verl} framework~\citep{verl-sheng2024hybridflow}, a flexible and efficient RL framework for LLM post-training. For RLVR, we use the default GRPO implementation in \texttt{verl} without modification, with the reward given by binary ground-truth verifier. For RL-Logit, we extend \texttt{verl} with a custom reward function that computes the scalar reward from the policy logits on expert answers conditioned on the question and generated CoT tokens. To stabilize training and avoid vanishing or exploding rewards, we use two reward variants:
\begin{itemize}
    \item Log-probability variant: $\max(0.1 \times \log \pi_\theta(a^\star \mid q, z), -1.0)$
    \item Perplexity variant: $10.0 \times \exp( \text{mean}( \log \pi_\theta(a^\star \mid q, z)))$
\end{itemize} For RARO, we further modify the framework to (i) support rewards derived from critic rollouts instead of direct verifiers, and (ii) implement a replay buffer and mixed data pipeline that interleaves policy and critic rollouts for stable joint training of the policy and critic.

\paragraph{Compute Setup.}
Unless otherwise specified, all non-RL methods (SFT, Rationalization, and DPO) are trained on a single node with 8$\times$H100 GPUs, regardless of model size or reasoning token budget. RL-style methods are more compute intensive: we train RLVR, RL-Logit, and our method on 2 nodes with 8$\times$H100 GPUs each for the 1.5B and 3B models, and on 4 such nodes (32 H100 GPUs in total) for the 7B model.

\subsection{Hyperparameters} \label{sec:implementation_details_hparams}

We summarize the core optimization hyperparameters used for all methods in Tables~\ref{tab:hparams_supervised_rl} and~\ref{tab:hparams_ours}. Unless otherwise specified, these settings are shared across all tasks (Countdown, DeepMath, and Poetry Writing) and model sizes described in Section~\ref{sec:experiments}.

\begin{table*}[t]
  \caption{\textbf{Shared hyperparameters across ours and baselines.} SFT and Rationalization share the same AdamW optimizer setup, while DPO uses a different configuration. All three share the same cosine learning-rate schedule. RLVR, RL-Logit, and RARO share the same underlying GRPO setup as described in Section~\ref{sec:method}.}
  \label{tab:hparams_supervised_rl}
  \centering
  \small
  \begin{minipage}[t]{0.31\textwidth}
    \centering
    \textbf{SFT \& Rationalization}\\[2pt]
    \begin{tabular}{ll}
      \toprule
      Hparam & Value \\
      \midrule
      Epochs & $4$ \\
      Batch size & $8$ \\
      Optim & AdamW \\
      \quad LR & $1\times 10^{-5}$ \\
      \quad Weight decay & $0.02$ \\
      \quad Max grad.\ norm & $1.0$ \\
      LR Sched & Cosine \\
      \quad Warmup ratio & $0.05$ \\
      \quad Min LR ratio & $0.03$ \\
      \quad Num cycles & $0.5$ \\
      \bottomrule
    \end{tabular}
  \end{minipage}
  \hfill
  \begin{minipage}[t]{0.31\textwidth}
    \centering
    \textbf{Iterative DPO}\\[2pt]
    \begin{tabular}{ll}
      \toprule
      Hparam & Value \\
      \midrule
      Epochs & $1$ \\
      Batch size & $128$ \\
      Optim & AdamW \\
      \quad LR & $1\times 10^{-6}$ \\
      \quad Weight decay & 0.01 \\
      \quad Max grad.\ norm & 1.0 \\
      LR Sched & Cosine \\
      \quad Warmup ratio & 0.05 \\
      \quad Min LR ratio & 0.03 \\
      \quad Num cycles & 0.5 \\
      $\beta_{\text{DPO}}$ & $0.1$ \\
      \bottomrule
    \end{tabular}
  \end{minipage}
  \hfill
  \begin{minipage}[t]{0.31\textwidth}
    \centering
    \textbf{RLVR \& RL-Logit \& RARO}\\[2pt]
    \begin{tabular}{ll}
      \toprule
      Hparam & Value \\
      \midrule
      Rollout batch & $1024$ \\
      Rollout temp. & $1.0$ \\
      Group size & $16$ \\
      Optim & AdamW \\
      \quad LR & $1\times 10^{-6}$ \\
      \quad Weight decay & $0.01$ \\
      \quad Max grad.\ norm & $1.0$ \\
      Train batch & $256$ \\
      Clip ratio & $[0.2, 0.28]$ \\
      KL coeff. & $10^{-3}$ \\
      \bottomrule
    \end{tabular}
  \end{minipage}
\end{table*}

\begin{table*}[t]
  \caption{\textbf{Hyperparameters for our method.} We use the relativistic critic and shared-LLM training setup described in Section~\ref{sec:method}, with tie rewards $(\tau_{\text{pol}},\tau_{\text{crit}})$ and loss weights $(\lambda_{\text{pol}},\lambda_{\text{crit}})$ chosen to balance exploration and critic supervision for each task.}
  \label{tab:hparams_ours}
  \centering
  \small
  \begin{minipage}{0.31\textwidth}
    \centering
    \textbf{RARO (Countdown)}\\[2pt]
    \begin{tabular}{ll}
      \toprule
      Hparam & Value \\
      \midrule
      $\tau_{\text{pol}}$ & $0.6$ \\
      $\tau_{\text{crit}}$ & $0.55$ \\
      $\lambda_{\text{pol}}$ & $1/2$ \\
      $\lambda_{\text{crit}}$ & $1/2$ \\
      \bottomrule
    \end{tabular}
  \end{minipage}
  \hfill
  \begin{minipage}{0.31\textwidth}
    \centering
    \textbf{RARO (DeepMath)}\\[2pt]
    \begin{tabular}{ll}
      \toprule
      Hparam & Value \\
      \midrule
      $\tau_{\text{pol}}$ & $0.6$ \\
      $\tau_{\text{crit}}$ & $0.55$ \\
      $\lambda_{\text{pol}}$ & $1/9$ \\
      $\lambda_{\text{crit}}$ & $8/9$ \\
      \bottomrule
    \end{tabular}
  \end{minipage}
  \hfill
  \begin{minipage}{0.31\textwidth}
    \centering
    \textbf{RARO (Poetry Writing)}\\[2pt]
    \begin{tabular}{ll}
      \toprule
      Hparam & Value \\
      \midrule
      $\tau_{\text{pol}}$ & $0.6$ \\
      $\tau_{\text{crit}}$ & $0.5$ \\
      $\lambda_{\text{pol}}$ & $1/3$ \\
      $\lambda_{\text{crit}}$ & $2/3$ \\
      \bottomrule
    \end{tabular}
  \end{minipage}
\end{table*}

\FloatBarrier

\subsection{Test-Time Scaling Algorithm} \label{sec:implementation_details_tts}

Here, we provide additional details for our Test-Time Scaling (TTS) algorithm. As described in Algorithm~\ref{alg:tts_tournament}, we implement TTS via a single-elimination tournament. Given a pool of candidate responses $\mathcal{Y}$ generated by the policy, we iteratively pair them and use the learned critic $C_\phi$ to select the better response. To mitigate the variance in the critic's generated reasoning, for each pair of responses $(y_A, y_B)$, we sample the critic $K$ times and use a majority vote to determine the winner. We use $K=4$ for all our TTS experiments. Tables~\ref{tab:countdown_tts} and~\ref{tab:main_results_tts} present the full results of RARO with TTS compared to baselines with identical TTS settings.

\begin{algorithm}[H]
    \caption{Single-Elimination Tournament for Test-Time Scaling}
    \label{alg:tts_tournament}
    \begin{algorithmic}[1]
    \REQUIRE Prompt $x$, Candidates $\mathcal{Y}$, Critic $C_\phi$, Votes $K$
    \ENSURE Best response $y^*$
    \WHILE{$|\mathcal{Y}| > 1$}
        \STATE $\mathcal{Y}_{next} \leftarrow \emptyset$
        \FOR{$i = 1$ to $|\mathcal{Y}|$ step 2}
            \IF{$i == |\mathcal{Y}|$} 
            \STATE $\mathcal{Y}_{next}.\text{append}(\mathcal{Y}[i])$; \textbf{continue} 
            \ENDIF
            \STATE $y_A, y_B \leftarrow \mathcal{Y}[i], \mathcal{Y}[i+1]$
            \STATE $v_A \leftarrow \sum_{k=1}^K \mathbb{I}(C_\phi(\cdot|x, y_A, y_B) \text{ favors } A)$
            \STATE $\mathcal{Y}_{next}.\text{append}(v_A > K/2 \ ? \ y_A : y_B)$
        \ENDFOR
        \STATE $\mathcal{Y} \leftarrow \mathcal{Y}_{next}$
    \ENDWHILE
    \STATE \textbf{return} $\mathcal{Y}[1]$
    \end{algorithmic}
    \end{algorithm}
  
\FloatBarrier

\section{Additional Results}

\subsection{Model Size Scaling on Countdown} \label{sec:countdown_scaling}

\begin{wraptable}{r}{0.45\textwidth}
    \vspace{-20pt}
    \centering
    \caption{Model Size Scaling on Countdown.}
    \label{tab:countdown_scaling}
    \begin{tabular}{lcc}
    \toprule
    Method & 1.5B & 3B \\
           & accuracy ($\%$) $\uparrow$ & accuracy ($\%$) $\uparrow$ \\
    \midrule
    RLVR & $57.7 \pm 1.6$ & $53.5 \pm 1.6$ \\
    RARO & $54.4 \pm 1.5$ & $49.7 \pm 1.6$ \\
    \bottomrule
    \end{tabular}
\end{wraptable}

\begin{figure}[h!]
    \centering
    \begin{minipage}{0.48\textwidth}
        \centering
        \includegraphics[width=\linewidth]{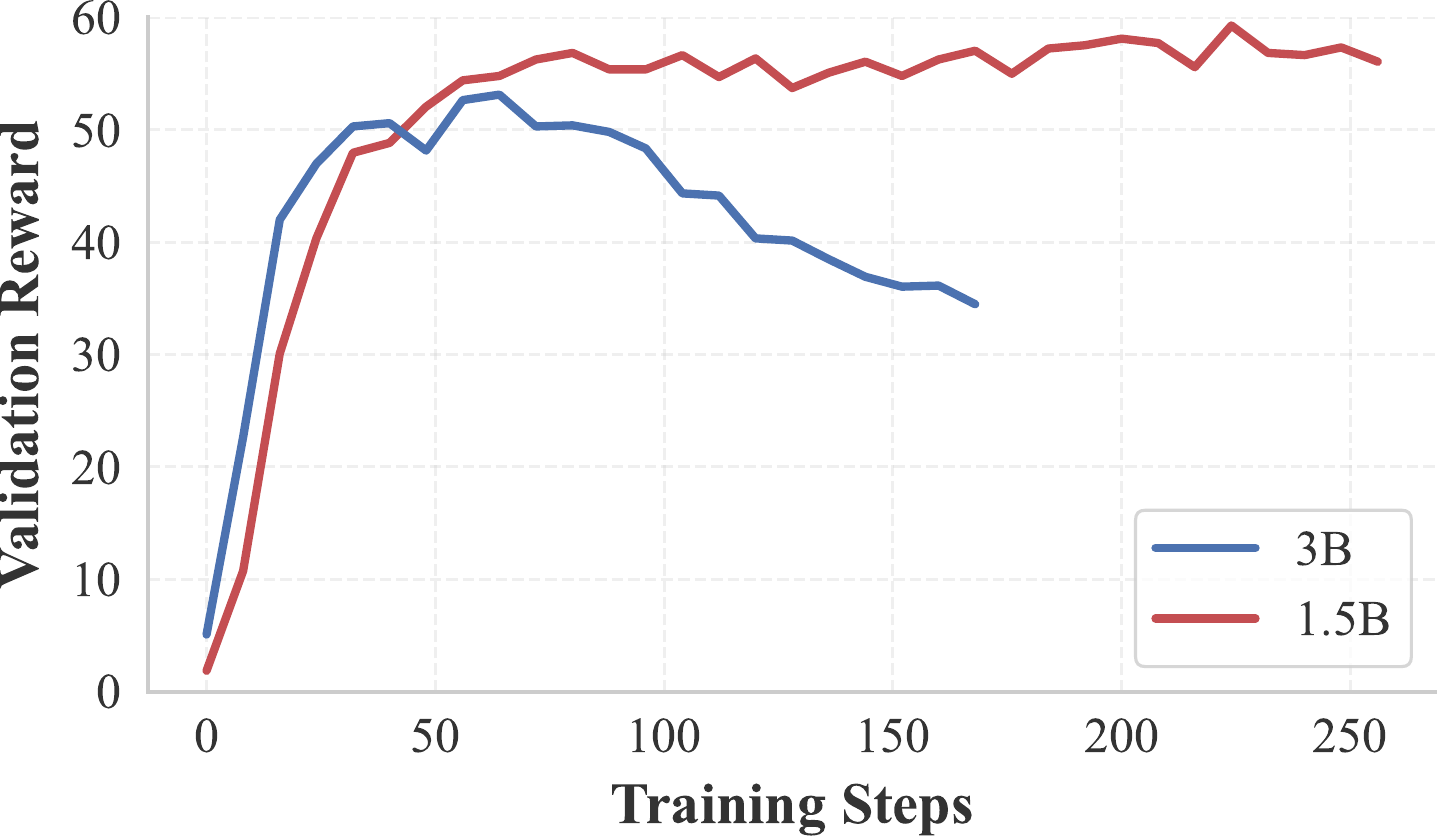}
        \caption{RLVR with a 3B model achieves lower performance than with a 1.5B model.}
        \label{fig:countdown_3b_verifiable}
    \end{minipage}
    \hfill
    \begin{minipage}{0.48\textwidth}
        \centering
        \includegraphics[width=\linewidth]{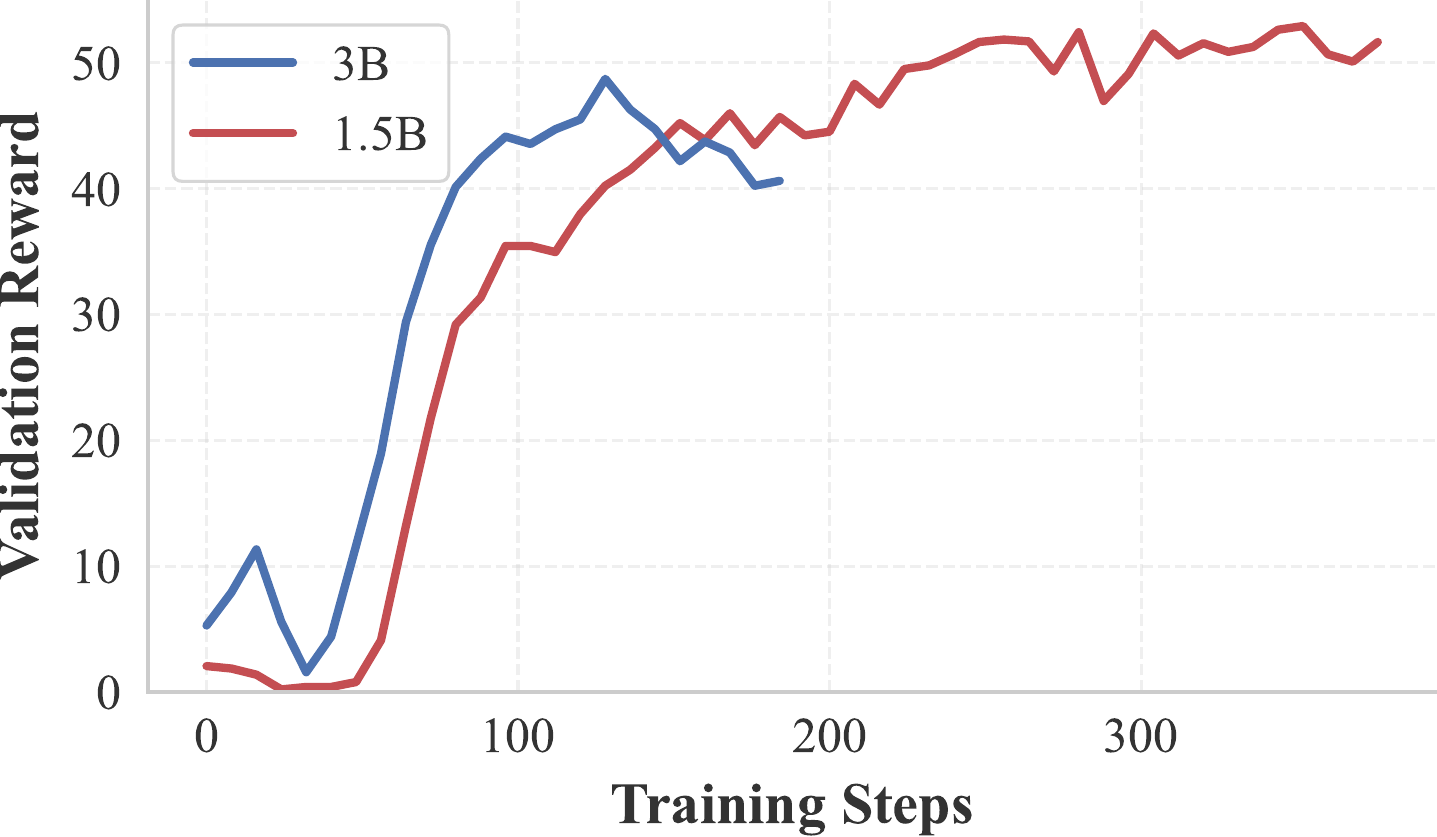}
        \caption{RARO similarly degrades from 1.5B to 3B.}
        \label{fig:countdown_3b}
    \end{minipage}
\end{figure}

We systematically study the effect of scaling model size on the performance on Countdown. In addition to the main results at 1.5B reported in Section~\ref{sec:results}, we conducted additional experiments at 3B. The verifiable baseline, RLVR, exhibits a performance regression, dropping from $57.7\%$ at 1.5B to $53.5\%$ at 3B (Figure~\ref{fig:countdown_3b_verifiable}). Similarly, we observe that the performance of RARO also degrades from $54.4\%$ to $49.7\%$ at 3B. Furthermore, as illustrated in Figure~\ref{fig:countdown_3b}, after initial improvements, both RLVR and RARO performance actively decreases as training progresses. While we do not have a definitive explanation, we hypothesize that larger models may be more prone to the training-inference log-probability mismatch problem~\citep{yao2025offpolicy}, leading to degradation when scaling model capacity. These results indicate that RARO does not inherently contribute to the performance plateau; rather, it is a systematic problem that we observe with RLVR as well. 

\subsection{Ablation Studies} \label{sec:ablation_studies}

\begin{figure}[t]
    \centering
    \begin{minipage}{0.48\textwidth}
        \centering
        \includegraphics[width=\linewidth]{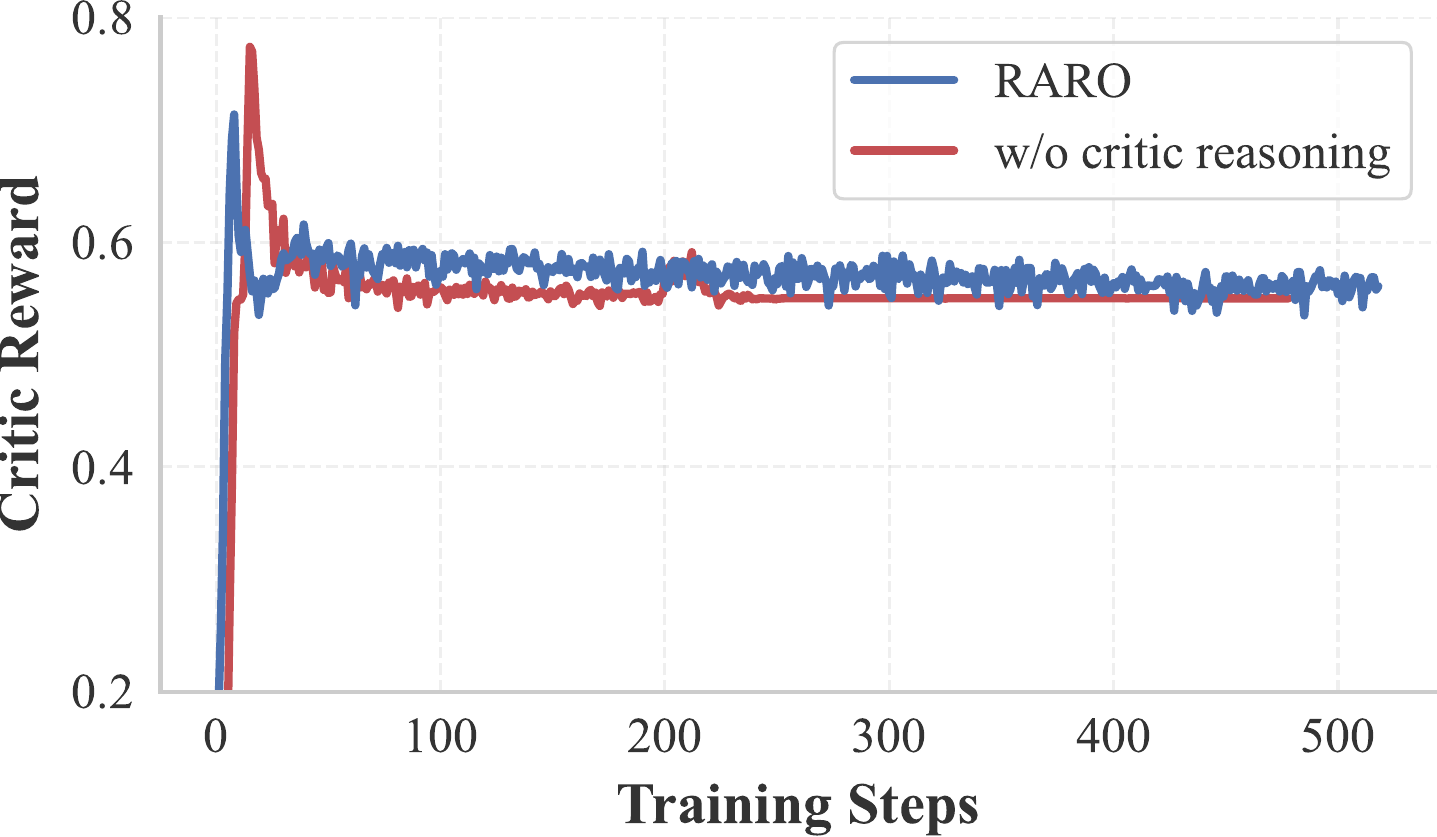}
        \caption{\textbf{No Critic Reasoning.} Without critic reasoning, the critic always outputs \texttt{tie}, preventing the policy from learning.}
        \label{fig:ablation_direct}
    \end{minipage}
    \hfill
    \begin{minipage}{0.48\textwidth}
        \centering
        \includegraphics[width=\linewidth]{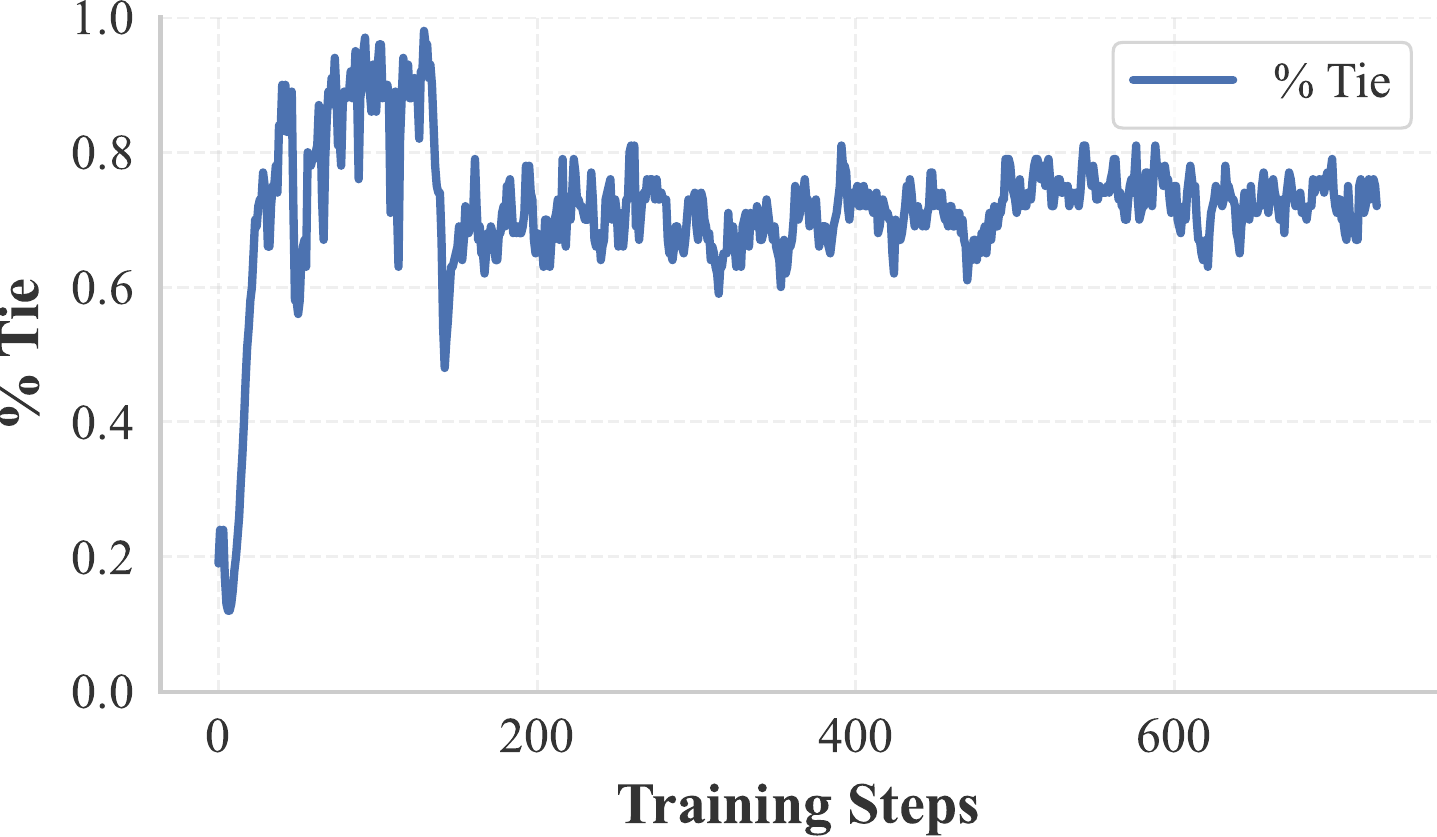}
        \caption{\textbf{Tie Distribution.} The critic learns to output \texttt{tie} stably for around $70\%$ of the outputs after around 150 training steps.}
        \label{fig:ablation_tie}
    \end{minipage}
\end{figure}

\paragraph{Necessity of Critic Reasoning.}
RARO relies on the critic performing explicit CoT reasoning before providing a final judgment. When this reasoning step is removed, the critic loses its capacity to make meaningful distinctions between responses. As shown in Figure~\ref{fig:ablation_direct}, instead of providing consistent signals, it collapses into a degenerate state, consistently outputting a \texttt{tie} response regardless of the quality of the policy or expert answer. This failure prevents the policy from receiving useful reward signals, stalling learning.

\begin{figure}[h]
    \centering
    \begin{minipage}{0.48\textwidth}
        \centering
        \includegraphics[width=\linewidth]{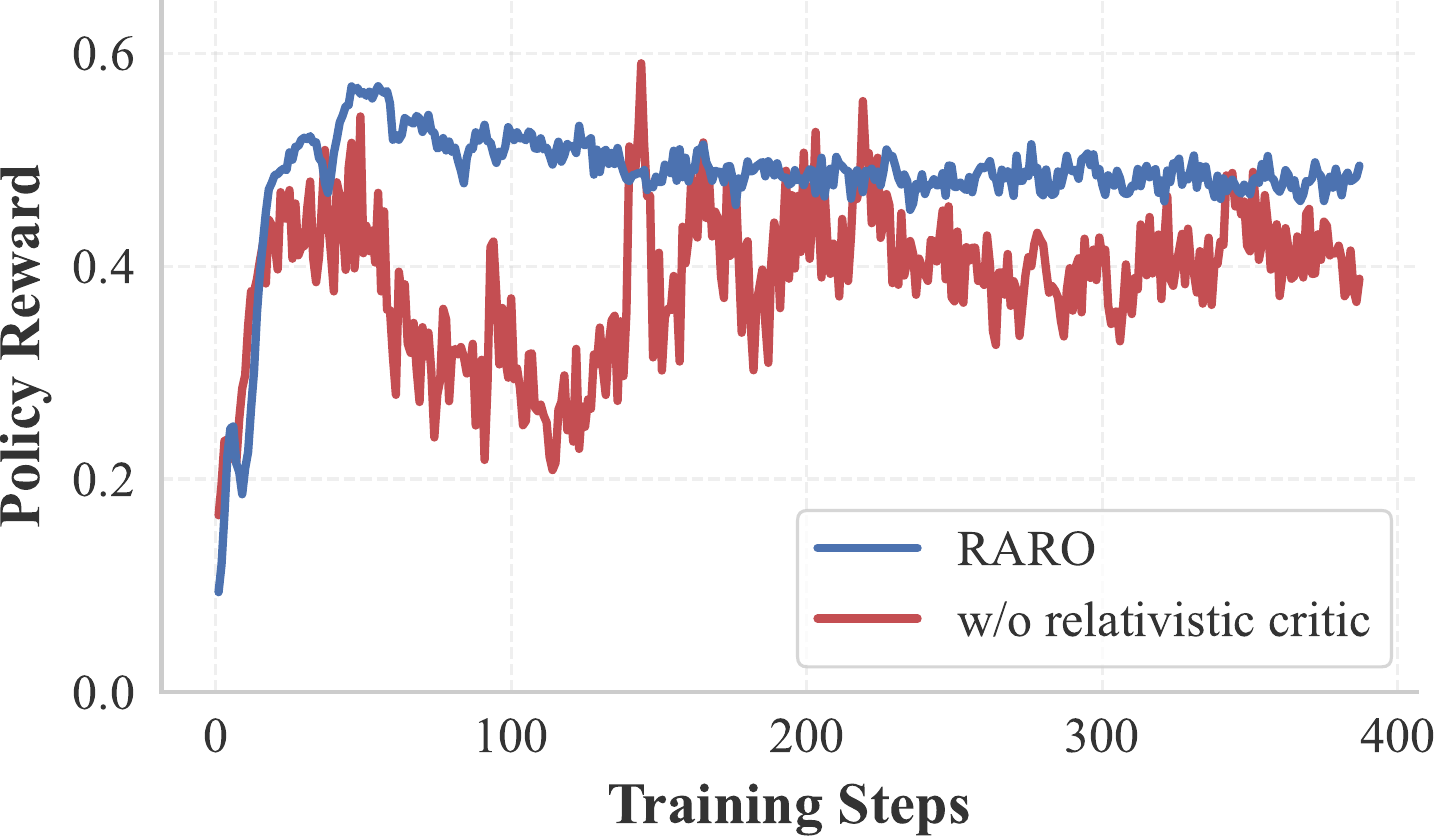}
        \caption{\textbf{No Relativistic Setup.} Without the relativistic setup, the policy reward during training exhibits high variance.}
        \label{fig:ablation_binary}
    \end{minipage}
    \hfill
    \begin{minipage}{0.48\textwidth}
        \centering
        \includegraphics[width=\linewidth]{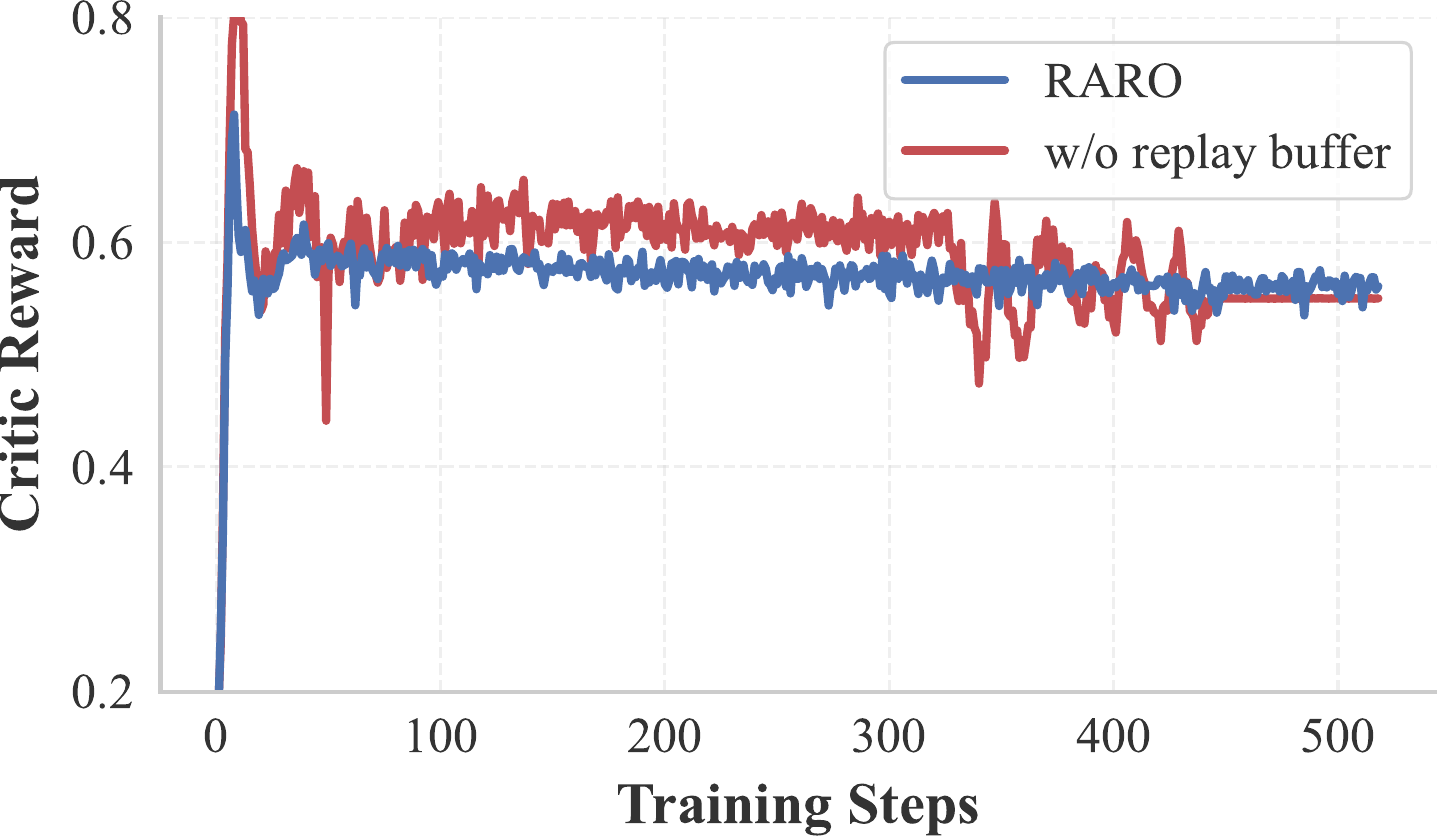}
        \caption{\textbf{No Replay Buffer.} Without a replay buffer, the training suffers from severe oscillations and eventual collapse.}
        \label{fig:ablation_norp}
    \end{minipage}
\end{figure}

\paragraph{Importance of Relativistic Setup.}
The relativistic critic evaluates the policy's answer and the expert's answer in a pairwise fashion rather than in isolation. Without this relativistic setup, the reward signal perceived by the policy exhibits significantly higher variance during training, as illustrated in Figure~\ref{fig:ablation_binary}. This instability suggests that the reference answer serves as a crucial anchor enabling stable optimization. We further demonstrate that the critic successfully learns to utilize the \texttt{tie} option defined in our relativistic setup. As shown in Figure~\ref{fig:ablation_tie}, the critic learns to output \texttt{tie} stably for around $70\%$ of the outputs after around 150 training steps. In addition, as shown in Table~\ref{tab:ablation_deepmath}, without the \texttt{tie} option, the final policy's performance drops from $41.3\%$ to $38.6\%$, indicating that the addition of the \texttt{tie} option contributes to the final policy performance.

\paragraph{Role of the Replay Buffer.}
Finally, the replay buffer is critical for preventing cycling dynamics. As shown in Figure~\ref{fig:ablation_norp}, removing the replay buffer causes the critic's training reward to oscillate severely after around 300 training steps. This suggests that the policy learns to exploit the critic's forgetfulness by cycling through adversarial patterns that temporarily fool the critic. This interaction eventually destabilizes the critic completely, leading it to default to a \texttt{tie} output, effectively halting progress.

\paragraph{Compute Overhead.}
Because RARO trains a policy and a critic, a natural concern is whether
it doubles the cost of standard RL. In our implementation, the policy
and critic share the same underlying LLM and are updated through a
mixed-batch GRPO step (Section~3.4), so the overhead is moderate rather
than $2\times$. Measured wall-clock step times are 310 s vs. 270 s on
Countdown and 330 s vs. 250 s on DeepMath against an RLVR baseline run
under identical hardware and rollout settings
(Table~\ref{tab:wall-clock-cost}). RARO therefore stays in the same
runtime regime as RLVR while removing the need for a task-specific
verifier; the larger cost gap is in sample efficiency, where RARO
requires more training iterations than RLVR because it must learn the
critic signal rather than receive oracle verifier feedback.

\begin{table}[t]
\centering
\caption{
Wall-clock training step time for RARO and RLVR under matched hardware
and rollout settings.
}
\label{tab:wall-clock-cost}
\begin{tabular}{lccc}
\toprule
Task & RLVR (s/step) & RARO (s/step) & Overhead \\
\midrule
Countdown & $270$ & $310$ & $+15\%$ \\
DeepMath  & $250$ & $330$ & $+32\%$ \\
\bottomrule
\end{tabular}
\end{table}
\FloatBarrier

\section{Additional Tables \& Figures} \label{sec:additional_tables_figures}

\begin{figure*}[h]
    \centering
    \scriptsize 
    \begin{minipage}[t]{0.32\textwidth}
        \begin{tcolorbox}[equal height group=inputs, colframe=blue!60!black, colback=white, title=Countdown, boxsep=2pt, left=2pt, right=2pt, top=2pt, bottom=2pt]
        Use the four numbers 19, 11, 48, and 6 to create an expression that equals 24.
        \end{tcolorbox}
        
        \begin{tcolorbox}[equal height group=reasoning, colframe=green!60!black, colback=white, title=RARO Policy, boxsep=2pt, left=2pt, right=2pt, top=2pt, bottom=2pt]
        To find an expression using the four numbers 19, 11, 48, and 6 that equals 24, let's try different combinations.
        \\ \\
        1. (48 - 11) * (19 - 6) = 37 * 13 = 481 (too high) \\
        2. (48 - 19) * (11 - 6) = 29 * 5 = 145 (too high)
        \\
        ...
        \\
        20. 48 / (19 - (11 + 6)) = 48 / (19 - 17) = 48 / 2 = 24 (correct)
        
        \tcblower
        \textbf{Answer: } \boxed{48 / (19 - (11 + 6))}
        \end{tcolorbox}
    \end{minipage}
    \hfill
    \begin{minipage}[t]{0.32\textwidth}
        \begin{tcolorbox}[equal height group=inputs, colframe=blue!60!black, colback=white, title=DeepMath, boxsep=2pt, left=2pt, right=2pt, top=2pt, bottom=2pt]
        Find the equation of the normal line to the curve \((x^2+y^2)^{3/2} = 2xy\) at the point \(\left( \frac{1}{\sqrt{2}}, \frac{1}{\sqrt{2}} \right)\).
        \end{tcolorbox}
        
        \begin{tcolorbox}[equal height group=reasoning, colframe=green!60!black, colback=white, title=RARO Policy, boxsep=2pt, left=2pt, right=2pt, top=2pt, bottom=2pt]
        To find the equation of the normal line to the curve \((x^2 + y^2)^{3/2} = 2xy\) at the point \(\left( \frac{1}{\sqrt{2}}, \frac{1}{\sqrt{2}} \right)\), we need to follow these steps:
        \\ \\
        1. Differentiate the curve implicitly to find the slope of the tangent line at the given point. \\
        2. Use the slope of the tangent line to find the slope of the normal line. \\
        3. Write the equation of the normal line using the point-slope form. \\
        \\ \\
        Let's start with...
        \tcblower
        \textbf{Answer: } \boxed{y=x}
        \end{tcolorbox}
    \end{minipage}
    \hfill
    \begin{minipage}[t]{0.32\textwidth}
        \begin{tcolorbox}[equal height group=inputs, colframe=blue!60!black, colback=white, title=Poetry Writing, boxsep=2pt, left=2pt, right=2pt, top=2pt, bottom=2pt]
        a poem about a child's fragmented memory of a father leaving, told in intimate, sensory free verse
        \end{tcolorbox}
        
        \begin{tcolorbox}[equal height group=reasoning, colframe=green!60!black, colback=white, title=RARO Policy, boxsep=2pt, left=2pt, right=2pt, top=2pt, bottom=2pt]
        To create such a poem, I'd:
        \\ \\
        1. Focus on sensory details that might linger in a child's memory.
        \\
        ...
        \\
        4. Incorporate physical and emotional sensations that capture the essence of the scene.
        \tcblower
        \textbf{Answer:}\\[2pt]
        One Memory of My Father Leaving\\
        \\
        There was the sound\\
        of a suitcase rolling\\
        on the hallway carpet;\\
        \ldots\\
        or a man thinking\\
        and trying not to cry.
        \end{tcolorbox}
    \end{minipage}
    \caption{\textbf{Qualitative examples for Countdown (1.5B), DeepMath (7B), and Poetry Writing (7B).} For each task, we show the input (top) and the truncated reasoning and answer sampled from policies trained with RARO (bottom). See Appendix~\ref{sec:additional_tables_figures} for full examples as well as example critic inputs and outputs.}
    \label{fig:qualitative_examples}
    \vspace{-100pt}
\end{figure*}

\FloatBarrier

\begin{table}[h]
    \caption{Tabular data for Countdown reasoning budget scaling results. Notably, result reported at a budget of 4096 tokens is derived from extrapolating test-time reasoning budget of the model trained at 2048 tokens.}
    \label{tab:budget_scale_data}
    \centering
    \begin{tabular}{lccccc}
    \toprule
    Budget & 256 & 512 & 1024 & 2048 & 4096 \\
    \midrule
    SFT & $40.7 \pm 1.6$ & $40.7 \pm 1.6$ & $40.7 \pm 1.6$ & $40.7 \pm 1.6$ & $40.7 \pm 1.6$ \\
    RARO & $33.1 \pm 1.5$ & $40.9 \pm 1.5$ & $51.2 \pm 1.6$ & $54.4 \pm 1.5$ & $61.3 \pm 1.5$ \\
    \bottomrule
    \end{tabular}
\end{table}

\FloatBarrier

\begin{table}[h]
    \centering
    \begin{minipage}[t]{0.48\textwidth}
        \centering
        \captionof{table}{RARO TTS scaling accuracy on DeepMath using Algorithm~\ref{alg:tts_tournament}}
        \label{tab:deepmath_tts_raro}
        \begin{tabular}{cccc}
        \toprule
        N & 1.5B & 3B & 7B \\
        \midrule
        1 & $41.6 \pm 1.9$ & $47.5 \pm 2.0$ & $57.8 \pm 1.9$ \\
        2 & $45.0 \pm 2.0$ & $51.7 \pm 2.0$ & $61.2 \pm 1.9$ \\
        4 & $45.1 \pm 2.0$ & $54.2 \pm 2.0$ & $63.4 \pm 2.0$ \\
        8 & $46.2 \pm 2.0$ & $54.8 \pm 2.0$ & $65.3 \pm 1.9$ \\
        16 & $53.6 \pm 2.3$ & $57.9 \pm 2.0$ & $68.4 \pm 2.0$ \\
        \bottomrule
        \end{tabular}
    \end{minipage}
    \hfill
    \begin{minipage}[t]{0.48\textwidth}
        \centering
        \captionof{table}{RLVR TTS scaling accuracy on DeepMath using best-of-N with oracle answers.}
        \label{tab:deepmath_tts_rlvr}
        \begin{tabular}{cccc}
        \toprule
        N & 1.5B & 3B & 7B \\
        \midrule
        1 & $50.9 \pm 1.9$ & $55.8 \pm 2.0$ & $66.2 \pm 2.0$ \\
        2 & $55.5 \pm 1.9$ & $63.4 \pm 1.9$ & $68.9 \pm 1.9$ \\
        4 & $59.6 \pm 1.9$ & $68.6 \pm 1.8$ & $69.8 \pm 1.9$ \\
        8 & $64.4 \pm 1.9$ & $72.5 \pm 1.7$ & $71.5 \pm 1.9$ \\
        16 & $66.1 \pm 1.8$ & $75.8 \pm 1.7$ & $76.9 \pm 1.9$ \\
        \bottomrule
        \end{tabular}
    \end{minipage}
\end{table}

\FloatBarrier

\begin{table}[h]
    \centering
    \begin{minipage}[t]{0.48\textwidth}
        \centering
        \captionof{table}{Complete results for Countdown at 1.5B.}
        \label{tab:countdown_all}
        \begin{tabular*}{\linewidth}{@{\extracolsep{\fill}}lc}
        \toprule
        Method & Countdown \\
        & accuracy ($\%$) $\uparrow$ \\
        \midrule
        RLVR (\emph{with verifier}) & $57.7 \pm 1.6$ \\
        \midrule
        Base & $2.0 \pm 0.4$ \\
        SFT & $40.7 \pm 1.6$ \\
        Rationalization & $12.5 \pm 1.0$ \\
        DPO & \\
        \qquad Round 1 & $40.4 \pm 1.5$ \\
        \qquad Round 2 & $32.5 \pm 1.4$ \\
        \qquad Round 3 & $32.2 \pm 1.4$ \\
        RL-Logit & $2.2 \pm 0.4$ \\
        \textbf{RARO} & $\mathbf{54.4 \pm 1.5}$ \\
        \bottomrule
        \end{tabular*}
    \end{minipage}
    \hfill
    \begin{minipage}[t]{0.48\textwidth}
        \centering
        \captionof{table}{Complete results for Countdown at 1.5B with TTS.}
        \label{tab:countdown_tts}
        \begin{tabular*}{\linewidth}{@{\extracolsep{\fill}}lc}
        \toprule
        Method & Countdown \\
        & accuracy ($\%$) $\uparrow$ \\
        \midrule
        RLVR (\emph{with verifier}) & $71.0 \pm 1.5$ \\
        \midrule
        Base & $4.2 \pm 0.6$ \\
        SFT & $42.4 \pm 1.6$ \\
        Rationalization & $11.2 \pm 1.0$ \\
        DPO & \\
        \qquad Round 1 & $43.1 \pm 1.6$ \\
        \qquad Round 2 & $34.8 \pm 1.5$ \\
        \qquad Round 3 & $31.6 \pm 1.4$ \\
        RL-Logit & $3.1 \pm 0.5$ \\
        \textbf{RARO} & $\mathbf{75.0 \pm 1.4}$ \\
        \bottomrule
        \end{tabular*}
    \end{minipage}
\end{table}

\FloatBarrier

\begin{table}[t]
    \caption{\textbf{Main results for DeepMath and Poetry.} We report the average and standard deviation of evaluation metrics for DeepMath and Poetry Writing across model scales with a reasoning token budget of $2048$. }
    \label{tab:main_results_all_appendix}
    \centering
    \begin{tabular}{lcccc}
    \toprule
    \multirow{2}{*}{Method} &
    \multicolumn{1}{c}{DeepMath} &
    \multicolumn{1}{c}{Poetry} &
    \multicolumn{1}{c}{Poetry} \\
    & accuracy ($\%$) $\uparrow$ & score (0-100) $\uparrow$ & \begin{tabular}[t]{@{}c@{}}win-rate ($\%$) $\uparrow$ \\ \emph{(vs. expert)} \end{tabular} \\
    \midrule
    \multicolumn{4}{l}{\textbf{1.5B}} \\
    \midrule
    \quad RLVR (\emph{with verifier}) &
    $50.9 \pm 1.9$ & \textsc{N/A} & \textsc{N/A} \\
    \midrule
    \quad Base & $29.6 \pm 1.9$ & $35.0 \pm 0.9$ & $0.0 \pm 0.0$ \\
    \quad SFT &
    $35.7 \pm 1.8$ & $53.7 \pm 1.0$ & $2.3 \pm 1.0$ \\
    \quad Rationalization &
    $34.5 \pm 2.0$ & $35.6 \pm 1.6$ & $0.8 \pm 0.5$ \\
    \quad DPO & & & \\
    \qquad Round 1 &
    $29.9 \pm 1.8$ & $48.6 \pm 0.9$ & $0.0 \pm 0.0$ \\
    \qquad Round 2 &
    $33.0 \pm 1.9$ & $10.3 \pm 0.5$ & $0.0 \pm 0.0$ \\
    \qquad Round 3 &
    $29.6 \pm 1.8$ & $29.3 \pm 1.0$ & $0.0 \pm 0.0$ \\
    \quad RL-Logit &
    $37.7 \pm 1.9$ & $36.4 \pm 0.7$ & $0.0 \pm 0.0$ \\
    \quad \textbf{RARO} &
    $\mathbf{41.3 \pm 1.9}$ & $\mathbf{67.8 \pm 0.8}$ & $\mathbf{7.8 \pm 1.7}$ \\
    \midrule
    \multicolumn{4}{l}{\textbf{3B}} \\
    \midrule
    \quad RLVR (\emph{with verifier}) &
    $55.8 \pm 2.0$ & \textsc{N/A} & \textsc{N/A} \\
    \midrule
    \quad Base &
    $39.4 \pm 1.9$ & $46.5 \pm 0.9$ & $0.0 \pm 0.0$ \\
    \quad SFT &
    $39.0 \pm 1.9$ & $57.4 \pm 1.0$ & $2.3 \pm 1.0$ \\
    \quad Rationalization &
    $32.3 \pm 1.9$ & $30.8 \pm 1.9$ & $0.4 \pm 0.4$ \\
    \quad DPO & & & \\
    \qquad Round 1 &
    $33.2 \pm 1.8$ & $58.7 \pm 0.9$ & $1.2 \pm 0.7$ \\
    \qquad Round 2 &
    $34.2 \pm 1.9$ & $57.1 \pm 1.0$ & $0.0 \pm 0.0$ \\
    \qquad Round 3 &
    $31.9 \pm 1.8$ & $69.8 \pm 0.8$ & $6.6 \pm 1.5$ \\
    \quad RL-Logit &
    $43.1 \pm 2.0$ & $46.9 \pm 0.8$ & $0.4 \pm 0.4$ \\
    \quad \textbf{RARO} &
    $\mathbf{49.1 \pm 1.9}$ & $\mathbf{71.9 \pm 0.8}$ & $\mathbf{17.2 \pm 2.4}$ \\
    \midrule
    \multicolumn{4}{l}{\textbf{7B}} \\
    \midrule
    \quad RLVR (\emph{with verifier}) &
    $66.2 \pm 1.9$ & \textsc{N/A} & \textsc{N/A} \\
    \midrule
    \quad Base &
    $44.2 \pm 2.1$ & $54.0 \pm 0.9$ & $1.2 \pm 0.7$ \\
    \quad SFT &
    $42.3 \pm 1.9$ & $65.4 \pm 1.0$ & $5.9 \pm 1.4$ \\
    \quad Rationalization &
    $48.6 \pm 1.9$ & $57.7 \pm 1.2$ & $5.1 \pm 1.3$ \\
    \quad DPO & & & \\
    \qquad Round 1 &
    $36.9 \pm 2.0$ & $61.6 \pm 0.9$& $3.5 \pm 1.1$ \\
    \qquad Round 2 &
    $36.5 \pm 1.9$ & $66.5 \pm 0.9$ & $5.1 \pm 1.4$ \\
    \qquad Round 3 &
    $32.8 \pm 1.9$ & $54.1 \pm 1.6$ & $3.9 \pm 1.2$ \\
    \quad RL-Logit &
    $49.3 \pm 2.0$ & $55.4 \pm 0.8$ & $3.9 \pm 1.2$ \\
    \quad \textbf{RARO} &
    $\mathbf{57.5 \pm 2.0}$ & $\mathbf{77.3 \pm 0.8}$ & $\mathbf{25.0 \pm 2.6}$ \\
    \bottomrule
    \end{tabular} 
\end{table}

\FloatBarrier

\begin{table}[t!]
    \caption{\textbf{Main results for DeepMath and Poetry with TTS.} Comparison of RARO against baselines with identical Test-Time Scaling settings.}
    \label{tab:main_results_tts}
    \centering
    \begin{tabular}{lcccc}
    \toprule
    \multirow{2}{*}{Method} &
    \multicolumn{1}{c}{DeepMath} &
    \multicolumn{1}{c}{Poetry} &
    \multicolumn{1}{c}{Poetry} \\
    & accuracy ($\%$) $\uparrow$ & score (0-100) $\uparrow$ & \begin{tabular}[t]{@{}c@{}}win-rate ($\%$) $\uparrow$ \\ \emph{(vs. expert)} \end{tabular} \\
    \midrule
    \multicolumn{4}{l}{\textbf{1.5B}} \\
    \midrule
    \quad RLVR (\emph{with verifier}) &
    $59.7 \pm 2.3$ & \textsc{N/A} & \textsc{N/A} \\
    \midrule
    \quad Base & 
    $26.9 \pm 6.2$ & $36.4 \pm 0.7$ & $0.0 \pm 0.0$ \\
    \quad SFT &
    $37.3 \pm 1.9$ & $55.1 \pm 1.1$ & $1.6 \pm 0.8$ \\
    \quad Rationalization &
    $42.6 \pm 2.8$ & $41.2 \pm 1.5$ & $0.0 \pm 0.0$ \\
    \quad DPO & & & \\
    \qquad Round 1 &
    $31.7 \pm 1.9$ & $49.9 \pm 0.9$ & $0.0 \pm 0.0$ \\
    \qquad Round 2 &
    $34.0 \pm 1.9$ & $9.5 \pm 0.4$ & $0.0 \pm 0.0$ \\
    \qquad Round 3 &
    $30.4 \pm 1.9$ & $30.1 \pm 1.1$ & $0.0 \pm 0.0$ \\
    \quad RL-Logit &
    $41.3 \pm 2.0$ & $38.0 \pm 0.7$ & $0.0 \pm 0.0$ \\
    \quad \textbf{RARO} &
    $\mathbf{53.6 \pm 2.3}$ & $\mathbf{67.7 \pm 0.8}$ & $\mathbf{8.2 \pm 1.8}$ \\
    \midrule
    \multicolumn{4}{l}{\textbf{3B}} \\
    \midrule
    \quad RLVR (\emph{with verifier}) &
    $67.5 \pm 2.1$ & \textsc{N/A} & \textsc{N/A} \\
    \midrule
    \quad Base &
    $49.7 \pm 2.9$ & $50.8 \pm 0.7$ & $0.4 \pm 0.4$ \\
    \quad SFT &
    $39.0 \pm 2.0$ & $57.2 \pm 1.0$ & $1.3 \pm 0.7$ \\
    \quad Rationalization &
    $42.7 \pm 2.6$ & $50.2 \pm 1.4$ & $2.0 \pm 0.8$ \\
    \quad DPO & & & \\
    \qquad Round 1 &
    $34.6 \pm 2.0$ & $57.5 \pm 0.9$ & $2.1 \pm 0.9$ \\
    \qquad Round 2 &
    $35.7 \pm 1.9$ & $55.8 \pm 0.9$ & $1.6 \pm 0.7$ \\
    \qquad Round 3 &
    $34.2 \pm 1.9$ & $70.3 \pm 0.8$ & $9.0 \pm 1.8$ \\
    \quad RL-Logit &
    $44.0 \pm 2.1$ & $51.1 \pm 0.7$ & $0.4 \pm 0.4$\\
    \quad \textbf{RARO} &
    $\mathbf{57.9 \pm 2.0}$ & $\mathbf{74.1 \pm 0.8}$ & $\mathbf{21.9 \pm 2.6}$ \\
    \midrule
    \multicolumn{4}{l}{\textbf{7B}} \\
    \midrule
    \quad RLVR (\emph{with verifier}) &
    $76.9 \pm 1.9$ & \textsc{N/A} & \textsc{N/A} \\
    \midrule
    \quad Base &
    $50.8 \pm 2.6$ & $58.8 \pm 0.6$ & $2.3 \pm 0.9$ \\
    \quad SFT &
    $42.1 \pm 2.0$ & $63.9 \pm 0.9$ & $4.7 \pm 1.4$ \\
    \quad Rationalization &
    $58.4 \pm 2.3$ & $55.1 \pm 1.2$ & $2.0 \pm 0.9$ \\
    \quad DPO & & & \\
    \qquad Round 1 &
    $42.9 \pm 2.3$ & $60.8 \pm 0.9$ & $3.5 \pm 1.1$ \\
    \qquad Round 2 &
    $36.8 \pm 1.9$ & $67.2 \pm 0.9$ & $4.3 \pm 1.3$ \\
    \qquad Round 3 &
    $42.9 \pm 2.3$ & $53.7 \pm 1.6$ & $2.7 \pm 1.0$ \\
    \quad RL-Logit &
    $51.0 \pm 2.1$ & $60.9 \pm 0.6$ & $1.6 \pm 0.8$ \\
    \quad \textbf{RARO} &
    $\mathbf{68.4 \pm 2.0}$ & $\mathbf{79.2 \pm 0.8}$ & $\mathbf{25.0 \pm 2.7}$ \\
    \bottomrule
    \end{tabular} 
\end{table}

\FloatBarrier

\begin{tcolorbox}[colframe=blue!60!black, colback=white, title=Countdown, boxsep=2pt, left=2pt, right=2pt, top=2pt, bottom=2pt]
Use the four numbers 19, 11, 48, and 6 to create an expression that equals 24.

\vspace{1.0em}

Remember:
\begin{itemize}
    \item Use each number exactly once
    \item You can only use +, -, *, / and parentheses
    \item The expression must be valid Python code that evaluates to 24
\end{itemize}

\vspace{1.0em}

Think through different combinations and show your reasoning. Once you find a valid solution, put your final answer inside \texttt{\textbackslash boxed\{\}} tags.

\vspace{1.0em}

For example, if the solution is (8 - 2) * 4 + 0, you would write: \texttt{\textbackslash boxed\{(8 - 2) * 4 + 0\}}

\vspace{1.0em}

The expression inside \texttt{\textbackslash boxed\{\}} must be directly evaluable in Python and should equal 24.
\end{tcolorbox}

\begin{tcolorbox}[colframe=green!60!black, colback=white, title=RARO Policy, boxsep=2pt, left=2pt, right=2pt, top=2pt, bottom=2pt]
\texttt{<think>} \\
To find an expression using the four numbers 19, 11, 48, and 6 that equals 24, let's try different combinations. \\

1. (48 - 11) * (19 - 6) = 37 * 13 = 481 (too high) \\
2. (48 - 19) * (11 - 6) = 29 * 5 = 145 (too high) \\
3. (48 * 11) / (19 - 6) = 528 / 13 = 40.615 (too high) \\
4. 48 / (19 - 11) + 6 = 48 / 8 + 6 = 6 + 6 = 12 (too low) \\
5. (48 - 11) * (19 / 6) = 37 * 3.167 = 115.167 (too high) \\
6. (48 / (19 - 11)) * 6 = (48 / 8) * 6 = 6 * 6 = 36 (too low) \\
7. (48 - 11) + (19 - 6) = 37 + 13 = 50 (too low) \\
8. 48 / (19 - 6) + 11 = 48 / 13 + 11 = 3.769 + 11 = 14.769 (too low) \\
9. 11 * (48 / (19 - 6)) = 11 * (48 / 13) = 11 * 3.769 = 41.459 (too low) \\
10. (48 / 6) + (19 - 11) = 8 + 8 = 16 (too low) \\
11. 19 - (48 / (11 - 6)) = 19 - (48 / 5) = 19 - 9.6 = 9.4 (too low) \\
12. (48 / (19 - 6)) + 11 = (48 / 13) + 11 = 3.769 + 11 = 14.769 (too low) \\
13. (48 - 11) + (19 - 6) = 37 + 13 = 50 (too low) \\
14. 19 - (48 / (11 + 6)) = 19 - (48 / 17) = 19 - 2.764 = 16.236 (too low) \\
15. (19 - 11) * (48 / 6) = 8 * 8 = 64 (too high) \\
16. (19 + 11) - (48 / 6) = 30 - 8 = 22 (too low) \\
17. 48 / (11 - (19 - 6)) = 48 / (11 - 13) = 48 / -2 = -24 (too low) \\
18. (48 / 6) + (19 - 11) = 8 + 8 = 16 (too low) \\
19. (48 / (19 - 6)) + 11 = (48 / 13) + 11 = 3.769 + 11 = 14.769 (too low) \\
20. 48 / (19 - (11 + 6)) = 48 / (19 - 17) = 48 / 2 = 24 (correct) \\

The final answer is: \texttt{\textbackslash boxed\{48 / (19 - (11 + 6))\}} \\
\texttt{</think>} \\
\boxed{48 / (19 - (11 + 6))}
\end{tcolorbox}
\captionof{figure}{Full qualitative example for the RARO policy for Countdown at 1.5B.}
\label{fig:qualitative_countdown_full}

\newpage

\begin{tcolorbox}[colframe=blue!60!black, colback=white, title=DeepMath, boxsep=2pt, left=2pt, right=2pt, top=2pt, bottom=2pt]
Find the equation of the normal line to the curve \((x^2+y^2)^{3/2} = 2xy\) at the point \(\left( \frac{1}{\sqrt{2}}, \frac{1}{\sqrt{2}} \right)\).
\end{tcolorbox}

\begin{tcolorbox}[colframe=green!60!black, colback=white, title=RARO Policy, boxsep=2pt, left=2pt, right=2pt, top=2pt, bottom=2pt, breakable]
\texttt{<think>} \\
To find the equation of the normal line to the curve \((x^2 + y^2)^{3/2} = 2xy\) at the point \(\left( \frac{1}{\sqrt{2}}, \frac{1}{\sqrt{2}} \right)\), we need to follow these steps:
\\ \\ 
1. Differentiate the curve implicitly to find the slope of the tangent line at the given point. \\
2. Use the slope of the tangent line to find the slope of the normal line. \\
3. Write the equation of the normal line using the point-slope form. \\
\\ \\
Let's start with implicit differentiation. Given the equation:
\[
(x^2 + y^2)^{3/2} = 2xy
\]
Differentiate both sides with respect to \(x\):
\[
\frac{d}{dx} \left( (x^2 + y^2)^{3/2} \right) = \frac{d}{dx} (2xy)
\]
Using the chain rule on the left side:
\[
\frac{3}{2} (x^2 + y^2)^{1/2} \cdot \frac{d}{dx} (x^2 + y^2) = 2y + 2x \frac{dy}{dx}
\]
Now, differentiate \(x^2 + y^2\) with respect to \(x\):
\[
\frac{3}{2} (x^2 + y^2)^{1/2} (2x + 2y \frac{dy}{dx}) = 2y + 2x \frac{dy}{dx}
\]
Simplify the left side:
\[
3 (x^2 + y^2)^{1/2} (x + y \frac{dy}{dx}) = 2y + 2x \frac{dy}{dx}
\]
Distribute and collect terms involving \(\frac{dy}{dx}\) on one side:
\[
3 (x^2 + y^2)^{1/2} x + 3 (x^2 + y^2)^{1/2} y \frac{dy}{dx} = 2y + 2x \frac{dy}{dx}
\]
\[
3 (x^2 + y^2)^{1/2} y \frac{dy}{dx} - 2x \frac{dy}{dx} = 2y - 3 (x^2 + y^2)^{1/2} x
\]
Factor out \(\frac{dy}{dx}\) on the left side:
\[
\frac{dy}{dx} \left( 3 (x^2 + y^2)^{1/2} y - 2x \right) = 2y - 3 (x^2 + y^2)^{1/2} x
\]
Solve for \(\frac{dy}{dx}\):
\[
\frac{dy}{dx} = \frac{2y - 3 (x^2 + y^2)^{1/2} x}{3 (x^2 + y^2)^{1/2} y - 2x}
\]

Next, we need to find the slope of the tangent line at the point \(\left( \frac{1}{\sqrt{2}}, \frac{1}{\sqrt{2}} \right)\):
Substitute \(x = \frac{1}{\sqrt{2}}\) and \(y = \frac{1}{\sqrt{2}}\) into the derivative:
\[
\frac{dy}{dx} = \frac{2 \left( \frac{1}{\sqrt{2}} \right) - 3 \left( \left( \left( \frac{1}{\sqrt{2}} \right)^2 + \left( \frac{1}{\sqrt{2}} \right)^2 \right) \right)^{1/2} \left( \frac{1}{\sqrt{2}} \right)}{3 \left( \left( \left( \frac{1}{\sqrt{2}} \right)^2 + \left( \frac{1}{\sqrt{2}} \right)^2 \right) \right)^{1/2} \left( \frac{1}{\sqrt{2}} \right) - 2 \left( \frac{1}{\sqrt{2}} \right)}
\]
Simplify the terms inside:
\[
x^2 + y^2 = \left( \frac{1}{\sqrt{2}} \right)^2 + \left( \frac{1}{\sqrt{2}} \right)^2 = \frac{1}{2} + \frac{1}{2} = 1
\]
So:
\[
\frac{dy}{dx} = \frac{2 \left( \frac{1}{\sqrt{2}} \right) - 3 \left( 1 \right)^{1/2} \left( \frac{1}{\sqrt{2}} \right)}{3 \left( 1 \right)^{1/2} \left( \frac{1}{\sqrt{2}} \right) - 2 \left( \frac{1}{\sqrt{2}} \right)}
= \frac{\frac{2}{\sqrt{2}} - \frac{3}{\sqrt{2}}}{\frac{3}{\sqrt{2}} - \frac{2}{\sqrt{2}}}
= \frac{\frac{2 - 3}{\sqrt{2}}}{\frac{3 - 2}{\sqrt{2}}} = \frac{-1}{1} = -1
\]
The slope of the tangent line at \(\left( \frac{1}{\sqrt{2}}, \frac{1}{\sqrt{2}} \right)\) is \(-1\). The slope of the normal line is the negative reciprocal of \(-1\), which is \(1\).

Now, use the point-slope form of the equation of a line \(y - y_1 = m (x - x_1)\), where \(m\) is the slope and \((x_1, y_1)\) is the point:
\[
y - \frac{1}{\sqrt{2}} = 1 \left( x - \frac{1}{\sqrt{2}} \right)
\]
Simplify:
\[
y - \frac{1}{\sqrt{2}} = x - \frac{1}{\sqrt{2}}
\]
\[
y = x
\]
Thus, the equation of the normal line is:
\[
\boxed{y = x}
\]
\texttt{</think>} \\
\boxed{y = x}
\end{tcolorbox}
\captionof{figure}{Full qualitative example for the RARO policy for DeepMath at 7B.}
\label{fig:qualitative_deepmath_full}

\newpage

\begin{tcolorbox}[colframe=blue!60!black, colback=white, title=Poetry Writing, boxsep=2pt, left=2pt, right=2pt, top=2pt, bottom=2pt]
a poem about a child's fragmented memory of a father leaving, told in intimate, sensory free verse

\vspace{1.0em}

Please carefully think about how to create an artistic poem while adhering to the prompt, and please enclose the final version of your poem (along with the title) in \texttt{<poem></poem>} tags
\end{tcolorbox}

\begin{tcolorbox}[colframe=green!60!black, colback=white, title=RARO Policy, boxsep=2pt, left=2pt, right=2pt, top=2pt, bottom=2pt]
\texttt{<think>} \\
To create such a poem, I'd: \\
1. Focus on sensory details that might linger in a child's memory. \\
2. Use free verse, with unrhymed, natural cadences to convey the fragmented and stream-of-consciousness nature of recollections. \\
3. Include elements that represent the end of a relationship, like a suitcase, luggage, or a car, without explicitly stating the scenario. \\
4. Incorporate physical and emotional sensations that capture the essence of the scene. \\
\texttt{</think>}

\texttt{<poem>} \\
One Memory of My Father Leaving \\

There was the sound \\
of a suitcase rolling \\
on the hallway carpet; \\

the thud of it \\
being set down in the sunroom \\
and a shadow of it carried away \\

by someone going somewhere. \\

It smelled of leather, or brown soap, \\
or something carried in it \\
-like salt air, or bare shoulders, \\

or a man thinking \\
and trying not to cry. \\
\texttt{</poem>}
\end{tcolorbox}
\captionof{figure}{Full qualitative example for the RARO policy for Poetry Writing at 7B.}
\label{fig:qualitative_poetry_full}

\newpage

\begin{tcolorbox}[colframe=blue!60!black, colback=white, title=Countdown, boxsep=2pt, left=2pt, right=2pt, top=2pt, bottom=2pt]
You are an expert mathematics critic whose goal is to compare two answers to a math problem and determine which one is correct.

\ldots 

Question: \\
\texttt{<question>} \\
Use the four numbers 20, 45, 10, and 23 to create an expression that equals 24.

\ldots

\texttt{</question>}

\vspace{1.0em}

Answer 1: \\
\texttt{<answer\_1>} \\
\boxed{(45-23)+(20/10)} \\
\texttt{</answer\_1>}

\vspace{1.0em}

Answer 2: \\
\texttt{<answer\_2>} \\
\boxed{(20+45*2)-23} \\
\texttt{</answer\_2>}
\end{tcolorbox}

\begin{tcolorbox}[colframe=green!60!black, colback=white, title=RARO Critic, boxsep=2pt, left=2pt, right=2pt, top=2pt, bottom=2pt]
\texttt{<think>} \\
To determine which answer is correct, let's carefully analyze both solutions.

For Answer 1: 
\boxed{(45-23)+(20/10)} \\
Let's follow these steps: \\
1. Perform the subtraction inside the first set of parentheses: \(45 - 23 = 22\) \\
2. Perform the division inside the second set of parentheses: \(20 / 10 = 2\) \\
3. Add the results: \(22 + 2 = 24\) \\

Answer 1 is correct. \\

For Answer 2: 
\boxed{(20+45*2)-23} \\
Let's follow these steps: \\
1. Perform the multiplication inside the parentheses: \(45 \times 2 = 90\) \\
2. Add the numbers inside the parentheses: \(20 + 90 = 110\) \\
3. Subtract 23: \(110 - 23 = 87\) \\

Answer 2 is incorrect. \\
\texttt{</think>} \\
\boxed{\text{Answer 1}}
\end{tcolorbox}
\captionof{figure}{Qualitative example of the RARO critic for Countdown at 1.5B.}
\label{fig:qualitative_countdown_critic_full}

\newpage

\begin{tcolorbox}[colframe=blue!60!black, colback=white, title=DeepMath, boxsep=2pt, left=2pt, right=2pt, top=2pt, bottom=2pt]
You are an expert mathematics critic whose goal is to compare two answers to a math problem and determine which one is correct.

\ldots

Question: \\
\texttt{<question>} \\
Evaluate the double integral \(\int \int _R x^2+y^2\,dx\,dy\) where \(R\) is the region bounded by the circle \(x^2 + y^2 = 4\) and the lines \(y = 0\) and \(y = x\). \\
\texttt{</question>}

\vspace{1.0em}

Answer 1: \\
\texttt{<answer\_1>} \\
\boxed{-\frac{8\pi}{5}} \\
\texttt{</answer\_1>}

\vspace{1.0em}

Answer 2: \\
\texttt{<answer\_2>} \\
\boxed{\pi} \\
\texttt{</answer\_2>}
\end{tcolorbox}

\begin{tcolorbox}[colframe=green!60!black, colback=white, title=RARO Critic, boxsep=2pt, left=2pt, right=2pt, top=2pt, bottom=2pt]
\texttt{<think>} \\
Let's evaluate the double integral \(\iint_R x^2 + y^2 \, dx \, dy\) where \(R\) is the region bounded by the circle \(x^2 + y^2 = 4\) and the lines \(y = 0\) and \(y = x\). \\

To solve the integral, we can convert to polar coordinates. The circle \(x^2 + y^2 = 4\) has a radius of 2, so in polar coordinates, the region \(R\) is described by \(0 \leq r \leq 2\) and \(0 \leq \theta \leq \frac{\pi}{4}\) (considering the quadrant where the region is). The integrand in polar coordinates is \(x^2 + y^2 = r^2\). \\

The Jacobian of the transformation from Cartesian to polar coordinates is \(r\), so the integral becomes:
\[
\iint_R x^2 + y^2 \, dx \, dy = \int_{0}^{\frac{\pi}{4}} \int_{0}^{2} r^2 \cdot r \, dr \, d\theta = \int_{0}^{\frac{\pi}{4}} \int_{0}^{2} r^3 \, dr \, d\theta.
\]

First, we integrate with respect to \(r\):
\[
\int_{0}^{2} r^3 \, dr = \left[ \frac{r^4}{4} \right]_{0}^{2} = \frac{2^4}{4} - \frac{0^4}{4} = \frac{16}{4} = 4.
\]

Next, we integrate with respect to \(\theta\):
\[
\int_{0}^{\frac{\pi}{4}} 4 \, d\theta = 4 \left[ \theta \right]_{0}^{\dfrac{\pi}{4}} = 4 \left( \dfrac{\pi}{4} - 0 \right) = \pi.
\]

Therefore, the value of the double integral is \(\pi\). \\

So, the answer is \(\boxed{\pi}\). \\
\texttt{</think>} \\
\boxed{\text{Answer 2}}
\end{tcolorbox}
\captionof{figure}{Qualitative example of the RARO critic for DeepMath at 7B.}
\label{fig:qualitative_deepmath_critic_full}

\newpage

\begin{tcolorbox}[colframe=blue!60!black, colback=white, title=Poetry Writing, boxsep=2pt, left=2pt, right=2pt, top=2pt, bottom=2pt]
You are an expert poetry critic and professor of creative writing. Your task is to compare two poems that were both written in response to the same prompt.

\ldots

Prompt: \\
\texttt{<prompt>} \\
Write a reflective free-verse poem about remembering the dead and how death feels sudden whether in disasters like the Titanic or in ordinary life. \\
\texttt{</prompt>}

\vspace{1.0em}

Poem 1: \\
\texttt{<poem\_1>} \\
On Missing Someone in a Disaster \\
\ldots \\
\texttt{</poem\_1>}

\vspace{1.0em}

Poem 2: \\
\texttt{<poem\_2>} \\
The Titanic \\
\ldots \\
\texttt{</poem\_2>}
\end{tcolorbox}

\begin{tcolorbox}[colframe=green!60!black, colback=white, title=RARO Critic, boxsep=2pt, left=2pt, right=2pt, top=2pt, bottom=2pt]
\texttt{<think>} \\
To evaluate these poems, we need to consider each dimension to grasp the key elements and compare them effectively. The prompt is to reflect on the experience of remembering the dead and perceiving death in contrast to major disasters like the Titanic. This prompts us to focus on both human and naturalistic death, the impact of memory, and the honor and acceptance of grief. \\

Both poems satisfy the prompt by offering reflections on the human condition related to death. They both focus on encounters with those who were absent in disasters and how death intrudes on life. However, Poem 1 uses a speculative metaphor (not \enquote{The Titanic,} but an abstractus timeframe-filled poem) to avoid repetition of a specific disaster. Poem 2, on the other hand, employs detailed imagery and simile relating directly to the well-known disaster. Both require the same thoughtful consideration of the prompt's themes. \\

The differences in focus and tone help to distinguish these views. While a survivalist approach might imply a romantic view of hardiness and mortality, this perspective doesn't rule out emotional reflection. Both poems visit the liminal space of collective memory and societal assumptions about mourning, driven by personal memory and regret. \\

In terms of literary merit and craftsmanship, both poems employ powerful strategies: evocative language, specific details from natural or social settings (both following the prompt well). However, the execution and spatial division among lines in Poem 2 yields a more static and inventive feeling, without many explicit sensory references or rhyme schemes. \\

Poem 1 then suffers from non-explicit time markers and symbols to depict the changing circumstances of disaster and loss. Unlike Poem 2, which doesn't explicitly use tragedy but rather deals subjectively with grief and loss, Poem 1's merging of analogies to the tragedy and reality illustrates the suspension and celebration of the poem's thorough reflection on these serious ideas. \\
\texttt{</think>}

Final Judgment:

\boxed{\text{Poem 2}}
\end{tcolorbox}
\captionof{figure}{Qualitative example of the RARO critic for Poetry Writing at 7B.}
\label{fig:qualitative_poetry_critic_full}

\end{document}